\documentclass{article}

\usepackage{arxiv}

\usepackage[utf8]{inputenc}
\usepackage[T1]{fontenc}
\usepackage{hyperref}
\usepackage{url}
\usepackage{booktabs}
\usepackage{amsfonts}
\usepackage{amsmath}
\usepackage{nicefrac}
\usepackage{microtype}
\usepackage{graphicx}
\usepackage{natbib}
\usepackage{doi}
\usepackage{subcaption}
\usepackage{algorithm}
\usepackage{algorithmic}
\usepackage{tikz}
\usepackage{pgfplots}
\usepackage{xcolor}
\usepackage{enumitem}
\usepackage{multirow}
\usepackage{array}
\usepackage{listings}
\usepackage{amsthm}
\usepackage{tcolorbox}

\lstdefinelanguage{JavaScript}{
    keywords={const, let, var, function, async, await, return, if, else, for, while, switch, case, break, continue, new, try, catch, throw, class, extends, import, export, from, default, typeof, instanceof},
    morecomment=[l]{//},
    morecomment=[s]{/*}{*/},
    morestring=[b]',
    morestring=[b]",
    morestring=[b]`,
    sensitive=true
}

\lstdefinelanguage{Dockerfile}{
    keywords={FROM, WORKDIR, COPY, RUN, EXPOSE, ENV, USER, CMD, ENTRYPOINT, ARG, LABEL, ADD, VOLUME},
    morecomment=[l]{\#},
    morestring=[b]",
    sensitive=true
}

\newtheorem{theorem}{Theorem}[section]

\usetikzlibrary{shapes.geometric, arrows, positioning, fit, backgrounds, calc}
\pgfplotsset{compat=1.17}

\definecolor{primaryblue}{RGB}{59, 130, 246}
\definecolor{accentpurple}{RGB}{139, 92, 246}
\definecolor{successgreen}{RGB}{34, 197, 94}
\definecolor{primarygreen}{RGB}{34, 197, 94}
\definecolor{accentgreen}{RGB}{74, 222, 128}
\definecolor{primaryyellow}{RGB}{234, 179, 8}

\title{SYNTHOCR-GEN: A SYNTHETIC OCR DATASET GENERATOR FOR LOW-RESOURCE LANGUAGES- BREAKING THE DATA BARRIER}

\author{
    \textbf{*Haq Nawaz Malik} \\
    \texttt{orcid.org/0009-0003-1994-7640} \\
    \texttt{huggingface.co/Omarrran} \\
    \
    \And
    \textbf{Kh Mohmad Shafi} \\
    \texttt{orcid.org/0000-0002-4759-8412} \\
    \texttt{huggingface.co/mshafi710} \\
    \And
    \textbf{Tanveer Ahmad Reshi} \\
    \texttt{orcid.org/0009-0002-4312-361X} \\
    \texttt{huggingface.co/TanveerReshi} \\}

\hypersetup{
    pdftitle={SYNTHOCR-GEN: A SYNTHETIC OCR DATASET GENERATOR
FOR LOW-RESOURCE LANGUAGES- BREAKING THE DATA
BARRIER
},
    pdfsubject={Computer Vision, OCR, Low-Resource Languages, Dataset Generation},
    pdfauthor={Haq Nawaz Malik, Kh Mohmad Shafi, Tanveer Reshi},
    pdfkeywords={OCR, Optical Character Recognition, Low-Resource Languages, Kashmiri, Synthetic Data, Dataset Generation, Deep Learning},
}

\renewcommand{\shorttitle}{SYNTHOCR-GEN: Synthetic OCR Dataset Generator}

\begin{document}
\maketitle

\begin{abstract}
Optical Character Recognition (OCR) for low-resource languages remains a significant challenge due to the scarcity of large-scale annotated training datasets. Languages such as Kashmiri, with approximately 7 million speakers and a complex Perso-Arabic script featuring unique diacritical marks, currently lack support in major OCR systems including Tesseract, TrOCR, and PaddleOCR. Manual dataset creation for such languages is prohibitively expensive, time-consuming, and error-prone, often requiring word by word transcription of printed or handwritten text.

We present \textbf{SynthOCR-Gen}, an open-source synthetic OCR dataset generator specifically designed for low-resource languages. Our tool addresses the fundamental bottleneck in OCR development by transforming digital Unicode text corpora into ready-to-use training datasets. The system implements a comprehensive pipeline encompassing text segmentation (character, word, n-gram, sentence, and line levels), Unicode normalization with script purity enforcement, multi-font rendering with configurable distribution, and 25+ data augmentation techniques simulating real-world document degradations including rotation, blur, noise, and scanner artifacts.

Key innovations include: (1) a fully client-side browser-based architecture ensuring data privacy, (2) native support for right-to-left scripts with proper handling of Arabic-script combining characters and Kashmiri-specific diacritics, (3) seeded randomization for reproducible dataset generation, and (4) multi-format output compatible with CRNN, TrOCR, PaddleOCR, Tesseract, and HuggingFace ecosystems.

We demonstrate the efficacy of our approach by generating a 600,000-sample word-segmented Kashmiri OCR dataset, which we release publicly on HuggingFace. This work provides a practical pathway for bringing low-resource languages into the era of vision-language AI models, and the tool is openly available for researchers and practitioners working with underserved writing systems worldwide.
\end{abstract}


\section{Introduction}
\label{sec:introduction}

The rapid advancement of deep learning has revolutionized Optical Character Recognition (OCR), enabling near-human performance on text recognition tasks for well-resourced languages such as English, Chinese, and Arabic \citep{li2021trocr, du2020ppocr}. However, this progress has not extended equally to the world's approximately 7,000 languages, leaving a significant digital divide where speakers of low-resource languages cannot fully participate in the benefits of AI-powered text processing technologies.

\subsection{The Low-Resource Language Challenge}

Low-resource languages face a fundamental chicken-and-egg problem in OCR development: training modern deep learning models requires large-scale annotated datasets, yet creating such datasets manually for languages with limited digital presence is prohibitively expensive. Consider the typical workflow for creating an OCR dataset manually:

\begin{enumerate}[noitemsep]
    \item Collect printed documents or book scans in the target language
    \item Segment images into character level,  word-level or sentence level crops
    \item Manually transcribe each image segment character-by-character/word-by-word or sentence-by-sentence
    \item Verify transcriptions for accuracy
    \item Format data for the target OCR framework architecture model
\end{enumerate}

For a dataset of 100,000 samples, considered modest by modern deep learning standards, this process could require thousands of person-hours and remains highly susceptible to human error, particularly for scripts with complex diacritical systems.

\subsection{Kashmiri: A Case Study}

We focus on Kashmiri (ISO 639-3: \texttt{kas}) as an exemplary low-resource language. Kashmiri is an Indo-Aryan language spoken by approximately 7 million people, primarily in the Kashmir Valley of the Indian subcontinent. The language uses a modified Perso-Arabic script with several unique characteristics:
\begin{figure}[htbp]
    \centering
    \includegraphics[width=0.8\textwidth]{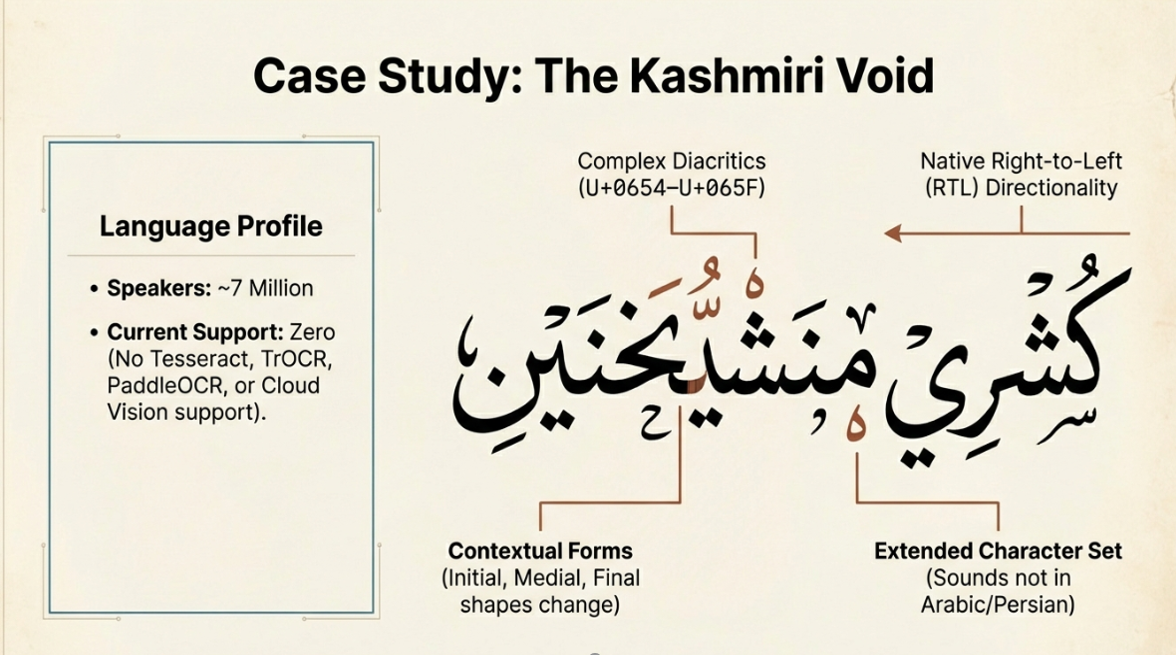}
    \caption{ Note: its an hypothetical visual (not real Kashmiri text) image text describing case study of Kashmiri language analysis for better understanding for international research readers.}
    \label{fig:case-kas-study}
\end{figure}

\begin{itemize}[noitemsep]
    \item \textbf{Extended Arabic character set}: Additional letters for sounds not present in Arabic or Persian
    \item \textbf{Complex diacritical system}: Extensive use of combining marks (U+0654--U+065F) for vowel representation
    \item \textbf{Right-to-left directionality}: Native RTL text flow with potential for embedded LTR numerals
    \item \textbf{Contextual letter forms}: Arabic-style initial, medial, final, and isolated letter shapes
\end{itemize}

Despite its significant speaker population and rich literary tradition, Kashmiri currently has \textit{zero} or negligence support in major OCR systems:

\begin{itemize}[noitemsep]
    \item \textbf{Tesseract}: No trained model available for Kashmiri script
    \item \textbf{TrOCR}: Pre-trained models do not include Kashmiri
    \item \textbf{PaddleOCR}: No Kashmiri language support
    \item \textbf{Google Cloud Vision}: Does not recognize Kashmiri text
    \item \textbf{Azure Computer Vision}: Kashmiri not among supported languages
    \item \textbf{Mistral OCR}: No support for Kashmiri language
    \item \textbf{DeepSeek OCR}: Does not support Kashmiri script
\end{itemize}

This complete absence of OCR capability prevents digitization of Kashmiri historical documents, limits accessibility tools for visually impaired Kashmiri speakers, and hinders the language's integration into modern AI pipelines.

\subsection{Synthetic Data: A Solution}

Synthetic data generation offers a compelling alternative to manual dataset curation. The core insight is straightforward: while collecting and transcribing real-world images is labor-intensive, the reverse process, \textbf{rendering known text into realistic images}, can be fully automated. Given a digital text corpus (which increasingly exists for many languages through literary archives, newspapers, and online sources), a synthetic data generator can produce unlimited training samples with perfect ground-truth annotations.

Previous work has demonstrated the effectiveness of synthetic data for OCR. The SynthText dataset \citep{gupta2016synthetic} and MJSynth (Synth90k) dataset \citep{jaderberg2014synthetic} enabled breakthrough performance on English scene text recognition. These works established that models trained on synthetic data can generalize effectively to real-world images when augmentation strategies adequately simulate natural image degradations.

\subsection{Contributions}

This paper makes the following contributions:

\begin{enumerate}
    \item \textbf{SynthOCR-Gen Tool}: We present an open-source, fully client-side synthetic OCR dataset generator designed specifically for low-resource languages. The tool supports multiple text segmentation modes, 25+ augmentation techniques, and outputs in formats compatible with all major OCR frameworks.
    
    \item \textbf{Kashmiri OCR Dataset}: We generate and publicly release a 600,000-sample word-segmented Kashmiri OCR dataset on HuggingFace, the first large-scale OCR dataset for this language.
    
    \item \textbf{Methodology for Low-Resource OCR}: We provide a comprehensive methodology that can be replicated for other low-resource languages, requiring only a Unicode text corpus and appropriate fonts.
    
    \item \textbf{Technical Solutions}: We address specific technical challenges including Kashmiri-specific diacritics preservation, RTL text rendering, and Unicode normalization for Persio-Arabic-script text.
\end{enumerate}

\subsection{Paper Structure Overview}

The remainder of this paper is organized as follows: Section~\ref{sec:related_work} reviews related work in synthetic data generation and low-resource language OCR. Section~\ref{sec:methodology} presents our system architecture and pipeline design. Section~\ref{sec:implementation} details implementation choices and technical solutions. Section~\ref{sec:Memory_Management_and_Sample_Storage} describes memory management and sample storage strategies. Section~\ref{sec:experiments} reports experiments on Kashmiri dataset generation. Section~\ref{sec:discussion} discusses implications, limitations, and broader applicability. Finally, Section~\ref{sec:conclusion} concludes with future research directions.

\section{Related Work}
\label{sec:related_work}

Our work builds upon three primary research areas: synthetic text image generation, OCR for right-to-left scripts, and low-resource language processing. We review key contributions in each area.

\subsection{Synthetic Text Image Datasets}

The use of synthetic data for training text recognition models was pioneered by \citet{jaderberg2014synthetic}, who introduced the MJSynth (Synth90k) dataset containing 9 million synthetically generated word images. This work demonstrated that convolutional neural networks trained exclusively on synthetic data could achieve competitive performance on real-world scene text benchmarks. The key insight was that sufficient variation in fonts, colors, backgrounds, and geometric transformations enables models to generalize beyond the synthetic training distribution.

\citet{gupta2016synthetic} extended this approach with SynthText, which rendered text onto natural scene backgrounds with realistic perspective transformations. Their contribution included a sophisticated text placement algorithm that identified suitable surface regions in background images, producing more visually realistic training data for scene text detection.

More recent work has explored domain-specific synthetic data generation. \citet{yim2021synthtiger} presented SynthTIGER, focusing on generating diverse text styles and complex backgrounds for robust text recognition training. Other researchers have explored 3D rendering engines and photorealistic scene synthesis for text image generation.

Our work differs from these approaches in its focus on \textit{document-style} text images for OCR rather than scene text, and specifically targets low-resource languages with complex scripts that have received limited attention in prior work.

\subsection{OCR for  Persio-Arabic-Script Languages}

Persio-Arabic script presents unique challenges for OCR systems due to its cursive nature, context-dependent letter forms, and extensive use of diacritical marks. Early work by \citet{amin1998survey} surveyed Arabic OCR challenges, identifying segmentation of connected characters as a primary obstacle.

The Tesseract OCR engine \citep{smith2007tesseract} supports Persio-Arabic through separately trained models, but quality varies significantly across  Persio-Arabic-script languages. While performance on Modern Standard Arabic has improved substantially with deep learning approaches, dialectal Arabic variants and related languages using Arabic script, including Persian, Urdu, and Kashmiri, often receive inadequate support due to limited training data availability.

Research on Urdu and Persian OCR has demonstrated the importance of language-specific datasets for achieving acceptable recognition accuracy. However, related languages using similar scripts, including Kashmiri, have not benefited from comparable research investment, leaving speakers of these languages without reliable OCR tools.

\subsection{Transformer-Based OCR}

The advent of vision-language transformers has significantly advanced OCR capabilities. \citet{li2021trocr} introduced TrOCR, combining a Vision Transformer (ViT) encoder with a text Transformer decoder, achieving state-of-the-art results on document text recognition benchmarks. The model's end-to-end architecture eliminates the need for explicit character segmentation, simplifying the recognition pipeline.

PaddleOCR \citep{du2020ppocr} provides a practical OCR toolkit supporting multiple languages and recognition approaches, including CRNN \citep{shi2017crnn} and attention-based models. The PP-OCR pipeline has been widely adopted for production OCR deployments due to its balance of accuracy and efficiency.

These architectural advances have significantly improved OCR quality for well-resourced languages but have not addressed the fundamental data scarcity problem for low-resource languages. State-of-the-art models require substantial training data, which remains unavailable for the majority of the world's writing systems.

\subsection{Low-Resource Language Processing}

The challenge of building NLP and vision systems for low-resource languages has attracted growing attention. \citet{joshi2020state} categorized the world's languages by their digital resource availability, finding that the vast majority fall into ``left-behind'' categories with minimal data for AI development. Their taxonomy highlighted the severe underrepresentation of languages outside the Indo-European family in NLP research.

\citet{rijhwani2020ocr} specifically addressed low-resource language OCR through multi-script transfer and post-correction techniques, demonstrating that recognition models can benefit from transfer across related scripts when character-level correspondences are established. Their work on post-OCR correction for endangered language texts showed promise for improving recognition quality with limited resources.

Cross-lingual transfer learning approaches have shown some success in text-based NLP for low-resource languages. However, these approaches are less directly applicable to OCR, where visual recognition of script-specific character shapes is paramount and cannot easily transfer across unrelated writing systems.

\subsection{Data Augmentation for OCR}

Effective data augmentation is critical for training robust OCR models. \citet{wigington2017data} systematically studied augmentation strategies for handwriting recognition, identifying geometric transformations, noise injection, and elastic distortions as particularly effective techniques for improving model generalization.

Research on document image enhancement has explored various degradation simulation techniques. \citet{souibgui2020synthetic} investigated combining real and synthetic data for historical document OCR, finding that augmented synthetic data improved recognition of degraded historical texts. Their work demonstrated the complementary value of synthetic data even when real annotated data is available.

Our augmentation pipeline incorporates insights from this literature \citep{wigington2017data, souibgui2020synthetic}, implementing a comprehensive set of 25+ transformations organized into geometric, blur, noise, degradation, and scanner effect categories. Critically, all augmentations preserve ground-truth labels, ensuring dataset integrity.

\subsection{Summary and Positioning}

While substantial progress has been made in synthetic data generation for English scene text and in Arabic-script OCR for major languages, a significant gap remains for low-resource languages. No prior work has specifically addressed the systematic generation of OCR training data for languages like Kashmiri that lack any existing model support. Our tool fills this gap by providing a principled, extensible framework that can be adapted to any language with available Unicode text and fonts.

\section{Methodology}
\label{sec:methodology}

This section presents the mathematical foundations, algorithmic design, and architectural principles of SynthOCR-Gen. We formalize each pipeline stage with precise definitions before describing the implementation.

\subsection{Notation Reference}

For reader convenience, Table~\ref{tab:notation} summarizes the key notation used throughout this section.

\begin{table}[htbp]
\centering
\caption{Summary of mathematical notation.}
\label{tab:notation}
\begin{tabular}{@{}cl@{}}
\toprule
\textbf{Symbol} & \textbf{Description} \\
\midrule
$\mathcal{C}$ & Input text corpus (sequence of Unicode characters) \\
$\mathcal{D}$ & Output dataset (set of image-label pairs) \\
$\mathcal{S}$ & Set of text segments after segmentation \\
$\mathcal{F}$ & Set of available fonts with probability weights \\
$I$ & Generated image (RGB tensor) \\
$y$ & Ground-truth text label \\
$N$ & Target dataset size (number of samples) \\
$H, W$ & Image height and width in pixels \\
$\Sigma$ & Alphabet (set of valid characters) \\
$\Theta$ & Configuration parameters \\
$\sigma, \psi, \rho, \phi, \pi$ & Pipeline stage operators \\
\bottomrule
\end{tabular}
\end{table}

\subsection{Complete System Workflow}

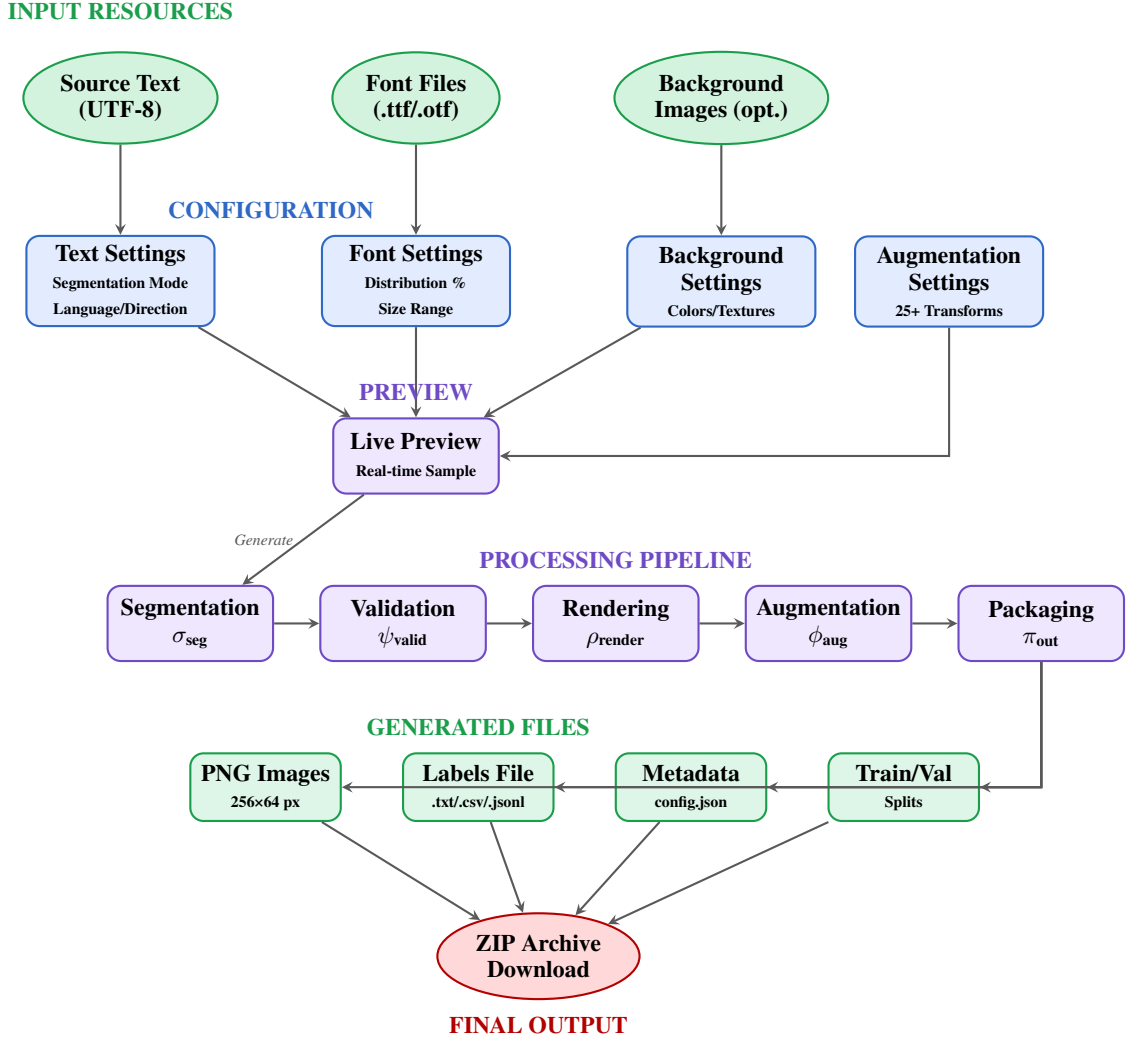
\begin{figure*}[htbp]
\centering
\begin{tikzpicture}[
    node distance=0.5cm and 0.8cm,
    input/.style={ellipse, draw=successgreen!80!black, fill=successgreen!20, thick,
                  minimum width=2.2cm, minimum height=0.9cm, align=center, font=\small\bfseries},
    config/.style={rectangle, draw=primaryblue!80!black, fill=primaryblue!15, thick,
                   minimum width=2.5cm, minimum height=1.2cm, align=center, 
                   font=\small\bfseries, rounded corners=5pt},
    process/.style={rectangle, draw=accentpurple!80!black, fill=accentpurple!15, thick,
                    minimum width=2.2cm, minimum height=1cm, align=center,
                    font=\small\bfseries, rounded corners=5pt},
    output/.style={rectangle, draw=successgreen!80!black, fill=successgreen!15, thick,
                   minimum width=2cm, minimum height=0.9cm, align=center,
                   font=\small\bfseries, rounded corners=5pt},
    final/.style={ellipse, draw=red!70!black, fill=red!15, thick,
                  minimum width=2.2cm, minimum height=0.9cm, align=center, font=\small\bfseries},
    arrow/.style={->, thick, >=stealth, color=gray!70!black},
    dasharrow/.style={->, thick, >=stealth, dashed, color=gray!50},
    label/.style={font=\tiny\itshape, color=gray!70!black}
]

\node[input] (text) {Source Text\\(UTF-8)};
\node[input, right=1.5cm of text] (fonts) {Font Files\\(.ttf/.otf)};
\node[input, right=1.5cm of fonts] (bgimages) {Background\\Images (opt.)};

\node[config, below=1.2cm of text] (textsettings) {Text Settings\\{\tiny Segmentation Mode}\\{\tiny Language/Direction}};
\node[config, below=1.2cm of fonts] (fontsettings) {Font Settings\\{\tiny Distribution \%}\\{\tiny Size Range}};
\node[config, below=1.2cm of bgimages] (bgsettings) {Background\\Settings\\{\tiny Colors/Textures}};

\node[config, right=0.5cm of bgsettings] (augsettings) {Augmentation\\Settings\\{\tiny 25+ Transforms}};

\node[process, below=1.2cm of fontsettings] (preview) {Live Preview\\{\tiny Real-time Sample}};

\node[process, below=1.2cm of preview, xshift=-3cm] (segment) {Segmentation\\$\sigma_{\text{seg}}$};
\node[process, right=0.6cm of segment] (validate) {Validation\\$\psi_{\text{valid}}$};
\node[process, right=0.6cm of validate] (render) {Rendering\\$\rho_{\text{render}}$};
\node[process, right=0.6cm of render] (augment) {Augmentation\\$\phi_{\text{aug}}$};
\node[process, right=0.6cm of augment] (package) {Packaging\\$\pi_{\text{out}}$};

\node[output, below=1.2cm of segment, xshift=1cm] (images) {PNG Images\\{\tiny 256×64 px}};
\node[output, right=0.8cm of images] (labels) {Labels File\\{\tiny .txt/.csv/.jsonl}};
\node[output, right=0.8cm of labels] (metadata) {Metadata\\{\tiny config.json}};
\node[output, right=0.8cm of metadata] (splits) {Train/Val\\{\tiny Splits}};

\node[final, below=1.2cm of labels, xshift=0.8cm] (zip) {ZIP Archive\\Download};

\draw[arrow] (text) -- (textsettings);
\draw[arrow] (fonts) -- (fontsettings);
\draw[arrow] (bgimages) -- (bgsettings);

\draw[arrow] (textsettings) -- (preview);
\draw[arrow] (fontsettings) -- (preview);
\draw[arrow] (bgsettings) -- (preview);
\draw[arrow] (augsettings) |- (preview);

\draw[arrow] (preview) -- node[label, left] {Generate} (segment);

\draw[arrow] (segment) -- (validate);
\draw[arrow] (validate) -- (render);
\draw[arrow] (render) -- (augment);
\draw[arrow] (augment) -- (package);

\draw[arrow] (package) |- (images);
\draw[arrow] (package) |- (labels);
\draw[arrow] (package) |- (metadata);
\draw[arrow] (package) |- (splits);

\draw[arrow] (images) -- (zip);
\draw[arrow] (labels) -- (zip);
\draw[arrow] (metadata) -- (zip);
\draw[arrow] (splits) -- (zip);

\node[above=0.3cm of text, font=\footnotesize\bfseries, color=successgreen!80!black] {INPUT RESOURCES};
\node[above=0.1cm of textsettings, xshift=2cm, font=\footnotesize\bfseries, color=primaryblue!80!black] {CONFIGURATION};
\node[above=0.1cm of preview, font=\footnotesize\bfseries, color=accentpurple!80!black] {PREVIEW};
\node[above=0.1cm of render, font=\footnotesize\bfseries, color=accentpurple!80!black] {PROCESSING PIPELINE};
\node[above=0.1cm of labels, font=\footnotesize\bfseries, color=successgreen!80!black] {GENERATED FILES};
\node[below=0.1cm of zip, font=\footnotesize\bfseries, color=red!70!black] {FINAL OUTPUT};

\end{tikzpicture}
\caption{Complete SynthOCR-Gen workflow diagram. The system accepts three types of inputs (source text, fonts, optional background images), processes them through configuration stages (text, font, background, augmentation settings), provides real-time preview, executes the five-stage processing pipeline ($\sigma \to \psi \to \rho \to \phi \to \pi$), and produces a downloadable ZIP archive containing PNG images, label files, metadata, and train/validation splits.}
\label{fig:complete_workflow}
\end{figure*}

Figure~\ref{fig:complete_workflow} presents the complete workflow of SynthOCR-Gen, from initial input through all configuration stages to final output generation and download.

\subsection{Problem Formulation}

Let $\mathcal{C} = \{c_1, c_2, \ldots, c_n\}$ denote a text corpus of $n$ Unicode characters. Our objective is to construct a dataset $\mathcal{D} = \{(I_i, y_i)\}_{i=1}^{N}$ where:
\begin{itemize}[noitemsep]
    \item $I_i \in \mathbb{R}^{H \times W \times 3}$ is an RGB image of dimensions $H \times W$ (height $\times$ width $\times$ 3 color channels)
    \item $y_i \in \Sigma^*$ is the ground-truth text label over alphabet $\Sigma$ (where $\Sigma^*$ denotes strings of any length)
    \item $N$ is the target dataset size (total number of image-label pairs to generate)
\end{itemize}

The generation function $\mathcal{G}: \Sigma^* \times \Theta \rightarrow \mathbb{R}^{H \times W \times 3}$ maps text to images parameterized by configuration $\Theta = (\theta_f, \theta_b, \theta_a)$ where:
\begin{itemize}[noitemsep]
    \item $\theta_f$ = font parameters (font files, sizes, distribution weights)
    \item $\theta_b$ = background parameters (colors, images, styles)
    \item $\theta_a$ = augmentation parameters (transforms, intensities, probabilities)
\end{itemize}

\subsection{System Architecture}

The system implements a pipeline architecture with five stages. Let the complete transformation be:

\begin{equation}
\mathcal{D} = \pi_{\text{out}} \circ \phi_{\text{aug}} \circ \rho_{\text{render}} \circ \psi_{\text{valid}} \circ \sigma_{\text{seg}}(\mathcal{C})
\label{eq:pipeline}
\end{equation}

\noindent where each operator represents a pipeline stage:
\begin{itemize}[noitemsep]
    \item $\sigma_{\text{seg}}$: \textbf{Segmentation} --- splits corpus into text segments
    \item $\psi_{\text{valid}}$: \textbf{Validation} --- filters and normalizes Unicode text
    \item $\rho_{\text{render}}$: \textbf{Rendering} --- converts text to images using fonts
    \item $\phi_{\text{aug}}$: \textbf{Augmentation} --- applies image transformations
    \item $\pi_{\text{out}}$: \textbf{Output} --- formats and packages the dataset
\end{itemize}

The symbol $\circ$ denotes function composition, meaning stages are applied sequentially from right to left.

\subsubsection{Pipeline Stage Diagram}

\begin{figure}[!htbp]
\centering
\begin{tikzpicture}[
    node distance=0.6cm,
    stage/.style={rectangle, draw=primaryblue, fill=primaryblue!15, thick, 
                  minimum width=2.2cm, minimum height=1cm, align=center, 
                  font=\small\bfseries, rounded corners=3pt},
    math/.style={rectangle, draw=accentpurple!70, fill=accentpurple!10, 
                 minimum width=1.8cm, minimum height=0.6cm, align=center, 
                 font=\footnotesize\itshape},
    arrow/.style={->, thick, >=stealth, primaryblue},
    data/.style={ellipse, draw=successgreen, fill=successgreen!15, 
                 minimum width=1.5cm, align=center, font=\small}
]

\node[data] (input) {$\mathcal{C}$\\Corpus};

\node[stage, right=of input] (seg) {Segmentation\\$\sigma_{\text{seg}}$};
\node[math, below=0.3cm of seg] (seg_math) {$\mathcal{C} \to \{s_j\}_{j=1}^M$};

\node[stage, right=of seg] (valid) {Validation\\$\psi_{\text{valid}}$};
\node[math, below=0.3cm of valid] (valid_math) {$s_j \to \tilde{s}_j$};

\node[stage, right=of valid] (render) {Rendering\\$\rho_{\text{render}}$};
\node[math, below=0.3cm of render] (render_math) {$\tilde{s}_j \to I_j^{(0)}$};

\node[stage, below=1.8cm of render] (aug) {Augmentation\\$\phi_{\text{aug}}$};
\node[math, below=0.3cm of aug] (aug_math) {$I_j^{(0)} \to I_j$};

\node[stage, left=of aug] (out) {Output\\$\pi_{\text{out}}$};
\node[math, below=0.3cm of out] (out_math) {$\{(I_j, y_j)\} \to \mathcal{D}$};

\node[data, left=of out] (output) {$\mathcal{D}$\\Dataset};

\draw[arrow] (input) -- (seg);
\draw[arrow] (seg) -- (valid);
\draw[arrow] (valid) -- (render);
\draw[arrow] (render) -- (aug);
\draw[arrow] (aug) -- (out);
\draw[arrow] (out) -- (output);

\end{tikzpicture}
\caption{Pipeline architecture with mathematical operators. Each stage transforms data according to the formalization shown below the stage boxes. Notation: $s_j$ = text segment, $\tilde{s}_j$ = validated segment, $I_j^{(0)}$ = raw image, $I_j$ = augmented image.}
\label{fig:pipeline}
\end{figure}
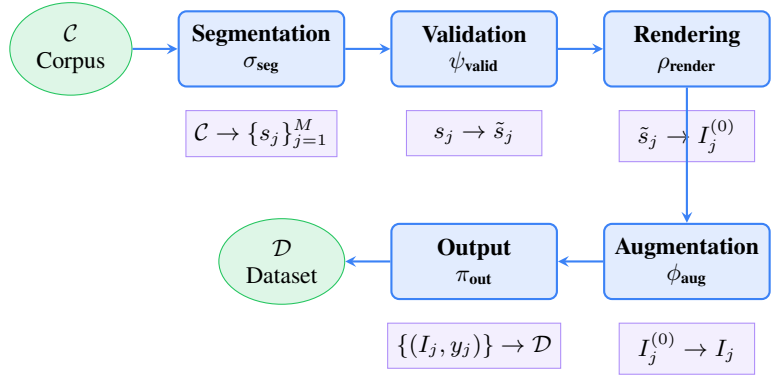

\subsection{Text Segmentation ($\sigma_{\text{seg}}$)}

\subsubsection{Formal Definition}

The segmentation operator partitions corpus $\mathcal{C}$ into a sequence of text segments:

\begin{equation}
\sigma_{\text{seg}}: \mathcal{C} \rightarrow \mathcal{S} = \{s_1, s_2, \ldots, s_M\}
\label{eq:segmentation}
\end{equation}

\noindent where:
\begin{itemize}[noitemsep]
    \item $\mathcal{C}$ = input text corpus (UTF-8 encoded string)
    \item $\mathcal{S}$ = output set of text segments
    \item $s_j$ = individual text segment (a substring of $\mathcal{C}$)
    \item $M$ = total number of segments produced
    \item $\Sigma^+$ = set of non-empty strings over alphabet $\Sigma$
\end{itemize}

Each segment $s_j \in \Sigma^+$ is a non-empty string. The partition depends on mode $m \in \{\text{char}, \text{word}, \text{ngram}, \text{sent}, \text{line}\}$.

\subsubsection{Segmentation Functions}

For each mode, we define the segmentation function:

\paragraph{Character Mode.} Using Unicode grapheme cluster boundaries:
\begin{equation}
\sigma_{\text{char}}(\mathcal{C}) = \{g_i : g_i \in \text{Graphemes}(\mathcal{C}), |g_i| > 0\}
\label{eq:char_seg}
\end{equation}

\noindent where:
\begin{itemize}[noitemsep]
    \item $g_i$ = a single grapheme cluster (visual character unit, may include combining marks)
    \item $\text{Graphemes}(\cdot)$ = function applying UAX \#29 grapheme cluster segmentation
    \item $|g_i|$ = length of grapheme in code points
\end{itemize}

\paragraph{Word Mode.} Using whitespace and punctuation delimiters:
\begin{equation}
\sigma_{\text{word}}(\mathcal{C}) = \text{Split}(\mathcal{C}, \mathcal{P}_{\text{delim}}) \setminus \{\epsilon\}
\label{eq:word_seg}
\end{equation}

\noindent where:
\begin{itemize}[noitemsep]
    \item $\text{Split}(\cdot, \cdot)$ = function that splits string by delimiter set
    \item $\mathcal{P}_{\text{delim}} = \{\text{space}, \text{U+060C}, \text{U+061B}, \text{!}, \ldots\}$ = delimiter set (including Arabic comma and semicolon)
    \item $\epsilon$ = empty string (excluded from result)
    \item $\setminus$ = set difference operator
\end{itemize}

\paragraph{N-gram Mode.} Sliding window over words:
\begin{equation}
\sigma_{\text{ngram}}(\mathcal{C}) = \bigcup_{n=2}^{4} \{w_i \circ w_{i+1} \circ \cdots \circ w_{i+n-1} : 1 \leq i \leq |W| - n + 1\}
\label{eq:ngram_seg}
\end{equation}

\noindent where:
\begin{itemize}[noitemsep]
    \item $W = \sigma_{\text{word}}(\mathcal{C})$ = word sequence from word-mode segmentation
    \item $w_i$ = the $i$-th word in sequence $W$
    \item $\circ$ = string concatenation with space separator
    \item $n$ = n-gram size (2 to 4 words per segment)
    \item $|W|$ = total number of words
    \item $\bigcup$ = union of all n-gram sets
\end{itemize}

\paragraph{Sentence Mode.} Using sentence-ending punctuation:
\begin{equation}
\sigma_{\text{sent}}(\mathcal{C}) = \text{Split}(\mathcal{C}, \{., ?, !, \text{U+061F}, \text{U+06D4}\}) \setminus \{\epsilon\}
\label{eq:sent_seg}
\end{equation}

\noindent where the delimiter set includes period (.), question marks (? and Arabic U+061F), exclamation (!), and Urdu full stop (U+06D4).

\paragraph{Line Mode.} Using line break characters:
\begin{equation}
\sigma_{\text{line}}(\mathcal{C}) = \text{Split}(\mathcal{C}, \{\backslash\text{n}, \backslash\text{r}\backslash\text{n}\}) \setminus \{\epsilon\}
\label{eq:line_seg}
\end{equation}

\noindent where $\backslash\text{n}$ = Unix newline, $\backslash\text{r}\backslash\text{n}$ = Windows newline.

\subsubsection{Length Filtering}

Segments are filtered by grapheme length:
\begin{equation}
\mathcal{S}' = \{s \in \mathcal{S} : \ell_{\min} \leq |\text{Graphemes}(s)| \leq \ell_{\max}\}
\label{eq:length_filter}
\end{equation}

\noindent where:
\begin{itemize}[noitemsep]
    \item $\mathcal{S}'$ = filtered segment set
    \item $\ell_{\min}$ = minimum allowed length (default: 1 grapheme)
    \item $\ell_{\max}$ = maximum allowed length (default: 50 graphemes)
    \item $|\text{Graphemes}(s)|$ = number of grapheme clusters in segment $s$
\end{itemize}

\subsection{Unicode Validation ($\psi_{\text{valid}}$)}

\subsubsection{Normalization}

Text undergoes Unicode normalization to canonical form:
\begin{equation}
\tilde{s} = \text{NFC}(s) = \text{Compose}(\text{Decompose}(s))
\label{eq:normalization}
\end{equation}

\noindent where:
\begin{itemize}[noitemsep]
    \item $s$ = input text segment
    \item $\tilde{s}$ = normalized text segment
    \item $\text{NFC}$ = Normalization Form Composed (Unicode standard)
    \item $\text{Decompose}(\cdot)$ = separates characters into base + combining marks
    \item $\text{Compose}(\cdot)$ = recombines into precomposed characters where possible
\end{itemize}

NFC ensures consistent character representation (e.g., ``é'' as single character vs. ``e'' + accent mark).

\subsubsection{Script Validation}

Let $\mathcal{U}_{\text{allowed}}$ be the set of allowed Unicode code point ranges. For Kashmiri:
\begin{align}
\mathcal{U}_{\text{Arabic}} &= [\texttt{0600}, \texttt{06FF}] \cup [\texttt{0750}, \texttt{077F}] \cup [\texttt{08A0}, \texttt{08FF}] \label{eq:arabic_range} \\
\mathcal{U}_{\text{Common}} &= [\texttt{0020}, \texttt{007F}] \cup [\texttt{2000}, \texttt{206F}] \label{eq:common_range} \\
\mathcal{U}_{\text{allowed}} &= \mathcal{U}_{\text{Arabic}} \cup \mathcal{U}_{\text{Common}} \label{eq:allowed_range}
\end{align}

\noindent where:
\begin{itemize}[noitemsep]
    \item $[\texttt{0600}, \texttt{06FF}]$ = Arabic block (main letters, diacritics)
    \item $[\texttt{0750}, \texttt{077F}]$ = Persio-Arabic kashmiri Supplement (additional letters)
    \item $[\texttt{08A0}, \texttt{08FF}]$ = Arabic Extended-A
    \item $[\texttt{0020}, \texttt{007F}]$ = Basic Latin (space, punctuation, digits)
    \item $[\texttt{2000}, \texttt{206F}]$ = General Punctuation
    \item $\cup$ = set union operator
\end{itemize}

The validation predicate:
\begin{equation}
\psi_{\text{valid}}(s) = 
\begin{cases}
\tilde{s} & \text{if } \forall c \in s: \text{codepoint}(c) \in \mathcal{U}_{\text{allowed}} \\
\bot & \text{otherwise (rejected)}
\end{cases}
\label{eq:validation}
\end{equation}

\noindent where:
\begin{itemize}[noitemsep]
    \item $\forall c \in s$ = for every character $c$ in segment $s$
    \item $\text{codepoint}(c)$ = Unicode code point value of character $c$
    \item $\bot$ = rejection symbol (segment is excluded from dataset)
\end{itemize}

\subsubsection{Diacritic Preservation}

Kashmiri diacritics in range $[\texttt{064B}, \texttt{065F}]$ are explicitly preserved:
\begin{equation}
\mathcal{D}_{\text{Kashmiri}} = \{\text{U+0654}, \text{U+0655}, \text{U+0656}, \text{U+0657}, \ldots\}
\label{eq:diacritics}
\end{equation}

\noindent where each code point represents a combining diacritical mark essential for Kashmiri vowel representation. These marks are not stripped during normalization.

\subsection{Image Rendering ($\rho_{\text{render}}$)}

\subsubsection{Font Selection}

Given font distribution $\mathcal{F} = \{(f_k, p_k)\}_{k=1}^{K}$ where $\sum_k p_k = 1$, font selection uses inverse transform sampling:
\begin{equation}
f^* = f_k \text{ where } k = \min\left\{j : \sum_{i=1}^{j} p_i \geq U\right\}, \quad U \sim \text{Uniform}(0, 1)
\label{eq:font_selection}
\end{equation}

\noindent where:
\begin{itemize}[noitemsep]
    \item $\mathcal{F}$ = font distribution set
    \item $f_k$ = the $k$-th font family (e.g., ``Noto Naskh Arabic'')
    \item $p_k$ = probability weight for font $f_k$ (e.g., 0.4 for 40\%)
    \item $K$ = total number of available fonts
    \item $f^*$ = selected font for current sample
    \item $U$ = uniform random variable in range $[0, 1)$
    \item $\sum_{i=1}^{j} p_i$ = cumulative probability up to font $j$
\end{itemize}

\subsubsection{Font Size Sampling}

Font size $z$ (in pixels) is sampled from configurable distribution:

\paragraph{Normal Distribution:}
\begin{equation}
z \sim \mathcal{N}\left(\mu = \frac{z_{\min} + z_{\max}}{2}, \sigma^2 = \left(\frac{z_{\max} - z_{\min}}{6}\right)^2\right)
\label{eq:normal_size}
\end{equation}

\noindent where:
\begin{itemize}[noitemsep]
    \item $z$ = font size in pixels
    \item $\mathcal{N}(\mu, \sigma^2)$ = normal distribution with mean $\mu$ and variance $\sigma^2$
    \item $z_{\min}, z_{\max}$ = minimum and maximum font sizes (e.g., 28px, 42px)
    \item $\mu$ = mean (midpoint of range)
    \item $\sigma$ = standard deviation (range/6 ensures 99.7\% of values within bounds)
\end{itemize}

Values are clipped to $[z_{\min}, z_{\max}]$ to ensure valid sizes.

\paragraph{Uniform Distribution:}
\begin{equation}
z \sim \text{Uniform}(z_{\min}, z_{\max})
\label{eq:uniform_size}
\end{equation}

\noindent where all sizes in range are equally likely.

\subsubsection{Canvas Rendering Model}

The rendering function produces an image:
\begin{equation}
I^{(0)} = \rho(s, f^*, z, \theta_b) \in \mathbb{R}^{H \times W \times 3}
\label{eq:render}
\end{equation}

\noindent where:
\begin{itemize}[noitemsep]
    \item $I^{(0)}$ = raw rendered image (before augmentation)
    \item $\rho$ = rendering function
    \item $s$ = text segment to render
    \item $f^*$ = selected font
    \item $z$ = font size
    \item $\theta_b$ = background parameters
    \item $\mathbb{R}^{H \times W \times 3}$ = 3D tensor of real values (RGB image)
\end{itemize}

The pixel value at position $(x, y)$ is computed as:
\begin{equation}
I^{(0)}(x, y) = 
\begin{cases}
c_{\text{text}} & \text{if } (x, y) \in \text{Glyph}(s, f^*, z) \\
c_{\text{bg}} & \text{otherwise}
\end{cases}
\label{eq:pixel}
\end{equation}

\noindent where:
\begin{itemize}[noitemsep]
    \item $I^{(0)}(x, y)$ = RGB color value at pixel coordinates $(x, y)$
    \item $c_{\text{text}} \in [0, 255]^3$ = text color (e.g., black = $(0, 0, 0)$)
    \item $c_{\text{bg}} \in [0, 255]^3$ = background color (e.g., white = $(255, 255, 255)$)
    \item $\text{Glyph}(s, f^*, z)$ = set of pixel coordinates covered by rendered text glyphs
\end{itemize}

\subsubsection{Text Positioning}

For RTL (right-to-left) text with alignment $a \in \{\text{left}, \text{center}, \text{right}\}$:
\begin{equation}
x_{\text{start}} = 
\begin{cases}
W - p_r - w_{\text{text}} & \text{if } a = \text{left} \land \text{RTL} \\
\frac{W - w_{\text{text}}}{2} & \text{if } a = \text{center} \\
p_l & \text{if } a = \text{right} \land \text{RTL}
\end{cases}
\label{eq:positioning}
\end{equation}

\noindent where:
\begin{itemize}[noitemsep]
    \item $x_{\text{start}}$ = horizontal starting position for text rendering
    \item $W$ = image width in pixels
    \item $w_{\text{text}}$ = rendered text width in pixels
    \item $p_l, p_r$ = left and right padding in pixels
    \item RTL = right-to-left script (Arabic, Kashmiri, etc.)
    \item $\land$ = logical AND operator
\end{itemize}

\subsection{Augmentation Pipeline ($\phi_{\text{aug}}$)}

\subsubsection{Probabilistic Application}

Each sample is augmented with probability $p_{\text{aug}}$:
\begin{equation}
I = 
\begin{cases}
\phi(I^{(0)}) & \text{with probability } p_{\text{aug}} \\
I^{(0)} & \text{with probability } 1 - p_{\text{aug}}
\end{cases}
\label{eq:aug_prob}
\end{equation}

\noindent where:
\begin{itemize}[noitemsep]
    \item $I$ = final output image
    \item $I^{(0)}$ = raw rendered image (no augmentation)
    \item $\phi$ = augmentation function (composition of transforms)
    \item $p_{\text{aug}}$ = augmentation probability (e.g., 0.7 for 70\%)
\end{itemize}

\subsubsection{Transform Composition}

The augmentation function composes selected transforms:
\begin{equation}
\phi = T_{k_m} \circ T_{k_{m-1}} \circ \cdots \circ T_{k_1}
\label{eq:transform_chain}
\end{equation}

\noindent where:
\begin{itemize}[noitemsep]
    \item $T_i$ = the $i$-th available transform (e.g., rotation, blur, noise)
    \item $\{k_1, \ldots, k_m\}$ = indices of selected transforms
    \item $m$ = number of transforms applied (satisfies $m \leq m_{\max}$)
    \item $m_{\max}$ = maximum transforms per sample (default: 4)
    \item $\circ$ = function composition (transforms applied right-to-left)
\end{itemize}

\subsubsection{Geometric Transforms}

\paragraph{Rotation.} Affine rotation by angle $\theta$:
\begin{equation}
T_{\text{rot}}(I) = I \circ R_\theta, \quad R_\theta = 
\begin{pmatrix}
\cos\theta & -\sin\theta & t_x \\
\sin\theta & \cos\theta & t_y \\
0 & 0 & 1
\end{pmatrix}
\label{eq:rotation}
\end{equation}

\noindent where:
\begin{itemize}[noitemsep]
    \item $R_\theta$ = $3 \times 3$ homogeneous rotation matrix
    \item $\theta$ = rotation angle in radians, sampled as $\theta \sim \text{Uniform}(-\theta_{\max}, \theta_{\max})$
    \item $\theta_{\max}$ = maximum rotation angle (default: 10° = 0.175 rad)
    \item $t_x, t_y$ = translation to keep image centered after rotation
    \item $\cos, \sin$ = trigonometric functions
\end{itemize}

\paragraph{Skew.} Shear transformation:
\begin{equation}
T_{\text{skew}}(I) = I \circ S, \quad S = 
\begin{pmatrix}
1 & s_x & 0 \\
s_y & 1 & 0 \\
0 & 0 & 1
\end{pmatrix}
\label{eq:skew}
\end{equation}

\noindent where:
\begin{itemize}[noitemsep]
    \item $S$ = $3 \times 3$ shear matrix
    \item $s_x$ = horizontal shear factor, $s_x \sim \text{Uniform}(-s_{\max}, s_{\max})$
    \item $s_y$ = vertical shear factor, $s_y \sim \text{Uniform}(-s_{\max}/2, s_{\max}/2)$
    \item $s_{\max}$ = maximum shear (default: 0.2)
\end{itemize}

\subsubsection{Blur Transforms}

\paragraph{Gaussian Blur.} Convolution with Gaussian kernel:
\begin{equation}
T_{\text{blur}}(I) = I * G_\sigma, \quad G_\sigma(x, y) = \frac{1}{2\pi\sigma^2} \exp\left(-\frac{x^2 + y^2}{2\sigma^2}\right)
\label{eq:gaussian_blur}
\end{equation}

\noindent where:
\begin{itemize}[noitemsep]
    \item $*$ = 2D convolution operator
    \item $G_\sigma$ = Gaussian kernel function
    \item $\sigma$ = standard deviation (blur strength), $\sigma \sim \text{Uniform}(\sigma_{\min}, \sigma_{\max})$
    \item $\sigma_{\min}, \sigma_{\max}$ = blur range (default: 0.5 to 2.0 pixels)
    \item $\exp$ = exponential function
    \item $(x, y)$ = kernel coordinates relative to center
\end{itemize}

\paragraph{Motion Blur.} Directional kernel simulating camera movement:
\begin{equation}
T_{\text{motion}}(I) = I * K_{\text{motion}}, \quad K_{\text{motion}}(i, j) = 
\begin{cases}
\frac{1}{k} & \text{if } j = \lfloor k/2 \rfloor \land |i - k/2| \leq k/2 \\
0 & \text{otherwise}
\end{cases}
\label{eq:motion_blur}
\end{equation}

\noindent where:
\begin{itemize}[noitemsep]
    \item $K_{\text{motion}}$ = linear blur kernel
    \item $k$ = kernel size (blur length in pixels)
    \item $(i, j)$ = kernel indices
    \item $\lfloor \cdot \rfloor$ = floor function
\end{itemize}
The kernel is rotated by angle $\alpha \sim \text{Uniform}(0, 2\pi)$ for random blur direction.

\subsubsection{Noise Injection}

\paragraph{Gaussian Noise.} Additive random noise:
\begin{equation}
T_{\text{noise}}(I)(x, y) = \text{clip}\left(I(x, y) + \eta, 0, 255\right), \quad \eta \sim \mathcal{N}(0, \sigma_n^2)
\label{eq:gaussian_noise}
\end{equation}

\noindent where:
\begin{itemize}[noitemsep]
    \item $\eta$ = noise value sampled from normal distribution
    \item $\sigma_n$ = noise standard deviation (intensity)
    \item $\mathcal{N}(0, \sigma_n^2)$ = zero-mean Gaussian with variance $\sigma_n^2$
    \item $\text{clip}(\cdot, 0, 255)$ = clamps values to valid pixel range
\end{itemize}

\paragraph{Salt-and-Pepper Noise.} Impulse noise simulating dust/scratches:
\begin{equation}
T_{\text{sp}}(I)(x, y) = 
\begin{cases}
0 & \text{with probability } p_{\text{sp}}/2 \text{ (pepper/black)} \\
255 & \text{with probability } p_{\text{sp}}/2 \text{ (salt/white)} \\
I(x, y) & \text{with probability } 1 - p_{\text{sp}} \text{ (unchanged)}
\end{cases}
\label{eq:salt_pepper}
\end{equation}

\noindent where $p_{\text{sp}}$ = total noise probability (default: 0.01 to 0.05).

\subsubsection{Degradation Effects}

\paragraph{JPEG Compression.} Lossy compression artifacts:
\begin{equation}
T_{\text{jpeg}}(I) = \text{Decode}_{\text{JPEG}}(\text{Encode}_{\text{JPEG}}(I, q))
\label{eq:jpeg}
\end{equation}

\noindent where:
\begin{itemize}[noitemsep]
    \item $\text{Encode}_{\text{JPEG}}(\cdot, q)$ = JPEG compression at quality level $q$
    \item $\text{Decode}_{\text{JPEG}}(\cdot)$ = JPEG decompression
    \item $q \sim \text{Uniform}(q_{\min}, q_{\max})$ = quality factor (default: 30 to 70)
    \item Lower $q$ = more visible compression artifacts
\end{itemize}

\paragraph{Resolution Degradation.} Downscale-upscale to simulate low-resolution capture:
\begin{equation}
T_{\text{res}}(I) = \text{Upsample}(\text{Downsample}(I, r), 1/r)
\label{eq:resolution}
\end{equation}

\noindent where:
\begin{itemize}[noitemsep]
    \item $\text{Downsample}(\cdot, r)$ = reduce image size by factor $r$
    \item $\text{Upsample}(\cdot, 1/r)$ = restore to original size
    \item $r \sim \text{Uniform}(r_{\min}, r_{\max})$ = scale factor (default: 0.3 to 0.7)
    \item Smaller $r$ = more pixelation/blur
\end{itemize}

\subsubsection{Augmentation Summary}

\begin{table}[htbp]
\centering
\caption{Complete augmentation transforms with mathematical formulation and default parameters.}
\label{tab:augmentation_math}
\begin{tabular}{@{}llll@{}}
\toprule
\textbf{Category} & \textbf{Transform} & \textbf{Formulation} & \textbf{Parameters} \\
\midrule
\multirow{2}{*}{Geometric} & Rotation & $R_\theta$ matrix (Eq.~\ref{eq:rotation}) & $\theta \in [-10^\circ, 10^\circ]$ \\
& Skew & $S$ shear matrix (Eq.~\ref{eq:skew}) & $s \in [-0.2, 0.2]$ \\
\midrule
\multirow{2}{*}{Blur} & Gaussian & $G_\sigma$ kernel (Eq.~\ref{eq:gaussian_blur}) & $\sigma \in [0.5, 2.0]$ \\
& Motion & Directional $K$ (Eq.~\ref{eq:motion_blur}) & $k \in [3, 7]$ pixels \\
\midrule
\multirow{2}{*}{Noise} & Gaussian & Additive $\mathcal{N}(0, \sigma^2)$ & $\sigma \in [5, 25]$ \\
& Salt-Pepper & Impulse (Eq.~\ref{eq:salt_pepper}) & $p \in [0.01, 0.05]$ \\
\midrule
\multirow{2}{*}{Degradation} & JPEG & Compress/decompress & $q \in [30, 70]$ \\
& Resolution & Downsample/upsample & $r \in [0.3, 0.7]$ \\
\midrule
\multirow{2}{*}{Lighting} & Brightness & Linear: $I' = I \cdot (1 + \Delta)$ & $\Delta \in [-0.15, 0.15]$ \\
& Contrast & Gamma: $I' = 255 \cdot (I/255)^\gamma$ & $\gamma \in [0.7, 1.3]$ \\
\bottomrule
\end{tabular}
\end{table}

\begin{figure}[htbp]
\centering
\includegraphics[width=0.9\linewidth]{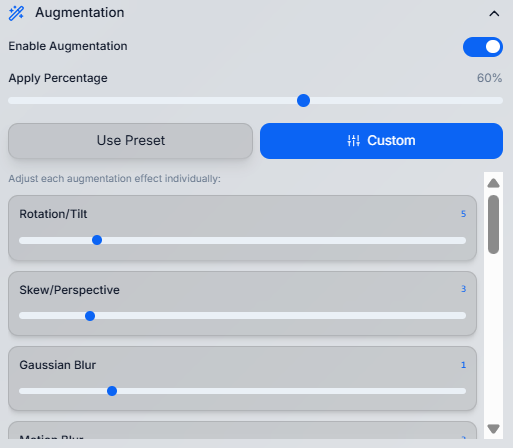}
\caption{Augmentation configuration interface in SynthOCR-Gen. Users can enable individual transforms, set intensity ranges, and control the probability of augmentation application. This granular control enables dataset customization for specific training requirements.}
\label{fig:augmentation_ui}
\end{figure}

Figure~\ref{fig:augmentation_ui} shows the augmentation configuration interface in SynthOCR-Gen, allowing users to enable/disable individual transforms and configure their intensity parameters.

\subsection{Seeded Randomization}

All stochastic operations use a seeded pseudo-random number generator (PRNG) for reproducibility.

\subsubsection{Linear Congruential Generator}

The PRNG follows the LCG recurrence relation:
\begin{equation}
X_{n+1} = (aX_n + c) \mod m
\label{eq:lcg}
\end{equation}

\noindent where:
\begin{itemize}[noitemsep]
    \item $X_n$ = current state (integer)
    \item $X_{n+1}$ = next state
    \item $a = 1103515245$ = multiplier constant
    \item $c = 12345$ = increment constant
    \item $m = 2^{31}$ = modulus
    \item $X_0$ = initial seed (user-configurable)
    \item $\mod$ = modulo operator
\end{itemize}

\subsubsection{Uniform Random Variate}

Uniform random values in $[0, 1)$ are derived as:
\begin{equation}
U_n = \frac{X_n}{m} \in [0, 1)
\label{eq:uniform_random}
\end{equation}

\noindent where $U_n$ can be used directly for probability comparisons or scaled to any range $[a, b]$ as $a + U_n \cdot (b - a)$.

\subsubsection{Reproducibility Theorem}

\begin{theorem}[Reproducibility]
Given identical inputs $(\mathcal{C}, \mathcal{F}, \Theta, X_0)$, the generation function $\mathcal{G}$ produces byte-identical output $\mathcal{D}$.
\end{theorem}

\begin{proof}
All random decisions (font selection, augmentation parameters, sample shuffling) are determined by the PRNG sequence $\{X_n\}_{n=0}^{\infty}$, which is uniquely determined by seed $X_0$ through recurrence (\ref{eq:lcg}). The deterministic rendering and encoding functions preserve this reproducibility.
\end{proof}

\begin{tcolorbox}[colback=gray!5!white, colframe=gray!75!black, title=\textbf{Implementation Note}]
We did not use an external seeded-RNG library. The project implements PRNGs directly in code: a linear congruential generator with $a = 1103515245$, $c = 12345$, $m = 2^{31}$ on the backend, while the web frontend uses a simple sin-based formula. Some places still fall back to the platform RNG (\texttt{Math.random}) when no seed is provided.
\end{tcolorbox}

\subsection{Output Format Adapters ($\pi_{\text{out}}$)}

The output operator serializes the dataset:
\begin{equation}
\pi_{\text{out}}: \{(I_j, y_j)\}_{j=1}^{N} \rightarrow (\mathcal{I}, \mathcal{L}, \mathcal{M})
\label{eq:output}
\end{equation}

\noindent where:
\begin{itemize}[noitemsep]
    \item $\{(I_j, y_j)\}_{j=1}^{N}$ = set of image-label pairs
    \item $\mathcal{I}$ = image archive (folder of PNG files)
    \item $\mathcal{L}$ = label file (format depends on target framework)
    \item $\mathcal{M}$ = metadata file (generation parameters, statistics)
\end{itemize}

\subsubsection{Format Specifications}

\begin{table}[htbp]
\centering
\caption{Output format specifications with file structures.}
\label{tab:output_formats_detail}
\begin{tabular}{@{}lp{6cm}l@{}}
\toprule
\textbf{Format} & \textbf{Label Structure} & \textbf{Framework} \\
\midrule
CRNN & \texttt{image\_000001.png\textbackslash t<text>} & PaddleOCR \\
TrOCR & \texttt{\{"image": "...", "text": "..."\}} & HuggingFace \\
CSV & \texttt{"images/...","<text>"} & General ML \\
HuggingFace & \texttt{file\_name,text} (metadata.csv) & HF Datasets \\
\bottomrule
\end{tabular}
\end{table}

\subsection{Complete Algorithm}

Algorithm~\ref{alg:generation} presents the complete generation procedure.

\begin{algorithm}[htbp]
\caption{SynthOCR-Gen Dataset Generation}
\label{alg:generation}
\begin{algorithmic}[1]
\REQUIRE Corpus $\mathcal{C}$, Configuration $\Theta$, Seed $X_0$, Target size $N$
\ENSURE Dataset $\mathcal{D} = \{(I_i, y_i)\}_{i=1}^{N}$

\STATE Initialize PRNG with seed $X_0$
\STATE $\mathcal{S} \leftarrow \sigma_{\text{seg}}(\mathcal{C})$ \COMMENT{Segment text into $M$ pieces}
\STATE $\mathcal{S}' \leftarrow \{\psi_{\text{valid}}(s) : s \in \mathcal{S}, \psi_{\text{valid}}(s) \neq \bot\}$ \COMMENT{Validate and filter}
\STATE Shuffle $\mathcal{S}'$ using PRNG

\FOR{$i = 1$ to $N$}
    \STATE $s \leftarrow \mathcal{S}'[i \mod |\mathcal{S}'|]$ \COMMENT{Select text (cycle if $N > M$)}
    \STATE $f^* \leftarrow \text{SelectFont}(\mathcal{F}, \text{PRNG})$ \COMMENT{Sample font by distribution}
    \STATE $z \leftarrow \text{SampleSize}(z_{\min}, z_{\max}, \text{PRNG})$ \COMMENT{Sample font size}
    \STATE $I^{(0)} \leftarrow \rho_{\text{render}}(s, f^*, z, \theta_b)$ \COMMENT{Render text to image}
    
    \IF{$\text{PRNG.next()} < p_{\text{aug}}$}
        \STATE $I_i \leftarrow \phi_{\text{aug}}(I^{(0)}, \text{PRNG})$ \COMMENT{Apply random augmentations}
    \ELSE
        \STATE $I_i \leftarrow I^{(0)}$ \COMMENT{Keep original image}
    \ENDIF
    
    \STATE $y_i \leftarrow s$ \COMMENT{Ground-truth label}
    \STATE Add $(I_i, y_i)$ to $\mathcal{D}$
\ENDFOR

\STATE $\mathcal{D} \leftarrow \pi_{\text{out}}(\mathcal{D})$ \COMMENT{Format and package output}
\RETURN $\mathcal{D}$
\end{algorithmic}
\end{algorithm}

\subsection{Complexity Analysis}

\subsubsection{Time Complexity}

Let $n = |\mathcal{C}|$ (corpus size), $M = |\mathcal{S}|$ (segment count), $N$ = target samples, $H \times W$ = image dimensions.

\begin{itemize}[noitemsep]
    \item \textbf{Segmentation}: $O(n)$ --- linear scan of corpus
    \item \textbf{Validation}: $O(M \cdot \bar{\ell})$ --- check each segment of average length $\bar{\ell}$
    \item \textbf{Rendering} (per sample): $O(H \cdot W)$ --- fill canvas pixels
    \item \textbf{Augmentation} (per sample): $O(H \cdot W \cdot k)$ --- apply $k$ transforms
    \item \textbf{Total}: $O(n + N \cdot H \cdot W \cdot k)$
\end{itemize}

\subsubsection{Space Complexity}

\begin{itemize}[noitemsep]
    \item \textbf{Corpus storage}: $O(n)$ --- input text
    \item \textbf{Segment storage}: $O(M \cdot \bar{\ell})$ --- extracted segments
    \item \textbf{Image buffer}: $O(H \cdot W)$ --- single canvas (reused)
    \item \textbf{Output archive}: $O(N \cdot H \cdot W / r)$ --- compressed images with ratio $r$
\end{itemize}

\section{Implementation Details}
\label{sec:implementation}

This section details the technical implementation of SynthOCR-Gen, presenting code structures, algorithmic optimizations, and platform-specific solutions.

\subsection{Technology Architecture}

\subsubsection{System Stack}

\begin{figure}[htbp]
\centering
\begin{tikzpicture}[
    node distance=1.2cm,
    layer/.style={rectangle, draw=primaryblue, fill=primaryblue!10, thick, minimum width=10cm, minimum height=1cm, align=center, font=\small},
    sublayer/.style={rectangle, draw=accentpurple!70, fill=accentpurple!5, minimum width=4.5cm, minimum height=0.7cm, align=center, font=\footnotesize},
    arrow/.style={->, thick, >=stealth}
]

\node[layer, fill=primaryblue!10] (ui) {User Interface Layer};
\node[layer, below=1.8cm of ui, fill=accentgreen!50] (app) {Application Logic Layer};
\node[layer, below=1.8cm of app, fill=primaryblue!50] (core) {Core Processing Layer};
\node[layer, below=1.8cm of core, fill=primaryyellow!10] (runtime) {Runtime Environment};

\node[sublayer, below=0.4cm of ui, xshift=-2.5cm, fill=primaryblue!10] (react) {React Components};
\node[sublayer, below=0.4cm of ui, xshift=2.5cm, fill=primaryblue!10] (tailwind) {Tailwind CSS};

\node[sublayer, below=0.4cm of app, xshift=-2.5cm, fill=primaryblue!10] (gen) {Generator Engine};
\node[sublayer, below=0.4cm of app, xshift=2.5cm, fill=primaryblue!10] (config) {Configuration Manager};

\node[sublayer, below=0.4cm of core, xshift=-2.5cm, fill=primaryblue!10] (canvas) {Canvas 2D API};
\node[sublayer, below=0.4cm of core, xshift=2.5cm, fill=primaryblue!10] (jszip) {JSZip Archiver};

\draw[arrow] (ui) -- (app);
\draw[arrow] (app) -- (core);
\draw[arrow] (core) -- (runtime);

\end{tikzpicture}
\caption{System Architecture with Color-Coded Layers}
\end{figure}
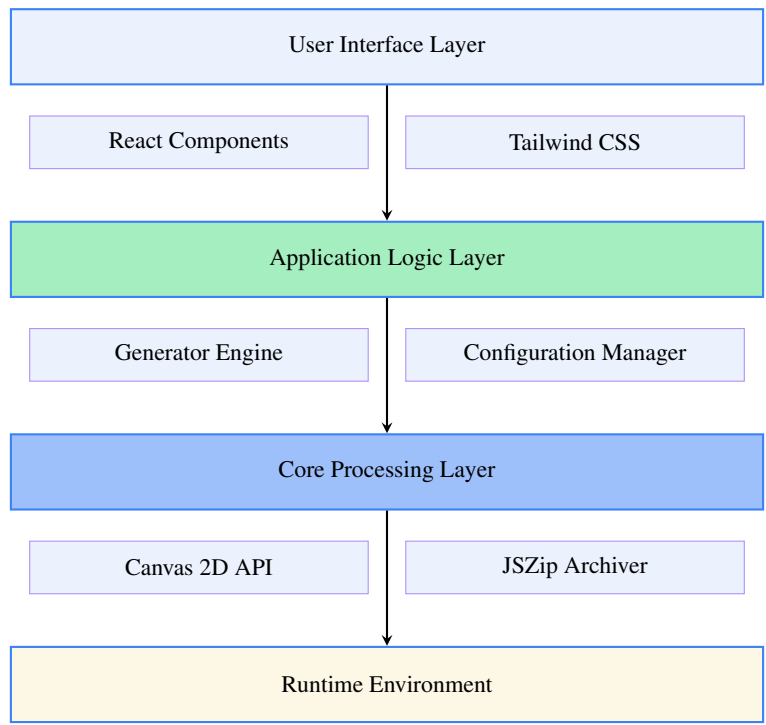

\begin{table}[htbp]
\centering
\caption{Technology stack components and versions.}
\label{tab:tech_stack}
\begin{tabular}{@{}lll@{}}
\toprule
\textbf{Layer} & \textbf{Technology} & \textbf{Version} \\
\midrule
Framework & Next.js & 14.x \\
Language & TypeScript & 5.x \\
UI Library & React & 18.x \\
Styling & Tailwind CSS & 3.x \\
Image Processing & Canvas 2D API & Native \\
Archive Creation & JSZip & 3.10.x \\
File Download & FileSaver.js & 2.0.x \\
Font Loading & FontFace API & Native \\
Deployment & Docker & 24.x \\
\bottomrule
\end{tabular}
\end{table}

\subsection{Unicode Processing Implementation}

\subsubsection{Grapheme Cluster Segmentation}

JavaScript strings use UTF-16 encoding, where a single visual character may span multiple code units. We employ the \texttt{Intl.Segmenter} API for correct grapheme handling:

\begin{lstlisting}[language=JavaScript, caption={Grapheme cluster segmentation for Arabic-script text.}, label={lst:grapheme}]
function segmentByGrapheme(text: string): string[] {
    const segmenter = new Intl.Segmenter('ar', { 
        granularity: 'grapheme' 
    });
    
    const segments: string[] = [];
    for (const { segment } of segmenter.segment(text)) {
        if (segment.trim().length > 0) {
            segments.push(segment);
        }
    }
    return segments;
}
\end{lstlisting}

The complexity of grapheme segmentation is $O(n)$ where $n$ is the string length.

\subsubsection{Script Range Validation}

Script validation uses Unicode code point ranges. The validation function:

\begin{lstlisting}[language=JavaScript, caption={Unicode script validation for Kashmiri.}, label={lst:unicode}]
const ARABIC_RANGES: [number, number][] = [
    [0x0600, 0x06FF],  // Arabic
    [0x0750, 0x077F],  // Arabic Supplement
    [0x08A0, 0x08FF],  // Arabic Extended-A
    [0xFB50, 0xFDFF],  // Presentation Forms-A
    [0xFE70, 0xFEFF],  // Presentation Forms-B
];

function isArabicScript(char: string): boolean {
    const cp = char.codePointAt(0);
    if (cp === undefined) return false;
    
    return ARABIC_RANGES.some(
        ([start, end]) => cp >= start && cp <= end
    );
}

function validateScript(text: string): boolean {
    for (const char of text) {
        if (!isArabicScript(char) && !isCommonChar(char)) {
            return false;
        }
    }
    return true;
}
\end{lstlisting}

\subsubsection{Normalization Process}

Unicode normalization ensures consistent character representation:

\begin{equation}
\text{NFC}(s) = \underbrace{\text{Compose}}_{\text{combine chars}} \circ \underbrace{\text{Decompose}}_{\text{split into base + marks}}(s)
\end{equation}

Implementation:

\begin{lstlisting}[language=JavaScript, caption={Unicode normalization.}, label={lst:normalize}]
function normalizeText(
    text: string, 
    form: 'NFC' | 'NFD' = 'NFC'
): string {
    return text.normalize(form);
}
\end{lstlisting}

\subsection{Seeded Random Number Generator}

\subsubsection{LCG Implementation}

The Linear Congruential Generator provides deterministic pseudo-randomness:

\begin{lstlisting}[language=JavaScript, caption={Seeded PRNG implementation.}, label={lst:prng}]
class SeededRandom {
    private seed: number;
    
    constructor(seed: number) {
        this.seed = seed;
    }
    
    next(): number {
        // LCG parameters (Numerical Recipes)
        this.seed = (this.seed * 1103515245 + 12345) % 2147483648;
        return this.seed / 2147483648;
    }
    
    range(min: number, max: number): number {
        return min + this.next() * (max - min);
    }
    
    int(min: number, max: number): number {
        return Math.floor(this.range(min, max + 1));
    }
    
    bool(p: number = 0.5): boolean {
        return this.next() < p;
    }
}
\end{lstlisting}

\subsubsection{Statistical Properties}

The LCG has period $m = 2^{31}$ and satisfies:
\begin{equation}
\mathbb{E}[X_n] = \frac{m-1}{2}, \quad \text{Var}(X_n) = \frac{(m-1)^2}{12}
\end{equation}

For our application, this provides sufficient randomness for dataset generation while maintaining reproducibility.

\subsection{Font Management System}

\subsubsection{Dynamic Font Loading}

Fonts are loaded asynchronously using the FontFace API:

\begin{lstlisting}[language=JavaScript, caption={Dynamic font loading from Data URL.}, label={lst:font}]
async function loadFont(
    name: string, 
    dataUrl: string
): Promise<string> {
    const familyName = `CustomFont_${name}_${Date.now()}`;
    
    try {
        const fontFace = new FontFace(
            familyName, 
            `url(${dataUrl})`
        );
        await fontFace.load();
        document.fonts.add(fontFace);
        return familyName;
    } catch (error) {
        console.warn(`Font load failed: ${name}`);
        return 'Arial';  // Fallback
    }
}
\end{lstlisting}

\subsubsection{Distribution-Based Selection}

Font selection follows the cumulative distribution function:

\begin{equation}
F(k) = P(K \leq k) = \sum_{i=1}^{k} p_i
\end{equation}

Implementation using inverse transform sampling:

\begin{lstlisting}[language=JavaScript, caption={Font selection by distribution.}, label={lst:fontselect}]
interface FontEntry {
    family: string;
    percentage: number;
}

function selectFont(
    fonts: FontEntry[], 
    rng: SeededRandom
): FontEntry {
    const u = rng.next() * 100;
    let cumulative = 0;
    
    for (const font of fonts) {
        cumulative += font.percentage;
        if (u < cumulative) {
            return font;
        }
    }
    return fonts[fonts.length - 1];
}
\end{lstlisting}

\subsection{Canvas Rendering Pipeline}

\subsubsection{Rendering Flow}

The rendering process follows a defined sequence:

\begin{figure}[htbp]
\centering
\begin{tikzpicture}[
    node distance=0.6cm,
    stepbox/.style={rectangle, draw=primaryblue, fill=primaryblue!10, thick,
                 minimum width=3cm, minimum height=0.7cm, align=center, font=\footnotesize},
    arrow/.style={->, thick, >=stealth}
]

\node[stepbox] (s1) {1. Create Canvas};
\node[stepbox, below=of s1] (s2) {2. Fill Background};
\node[stepbox, below=of s2] (s3) {3. Configure Context};
\node[stepbox, below=of s3] (s4) {4. Render Text};
\node[stepbox, below=of s4] (s5) {5. Export PNG Blob};

\draw[arrow] (s1) -- (s2);
\draw[arrow] (s2) -- (s3);
\draw[arrow] (s3) -- (s4);
\draw[arrow] (s4) -- (s5);

\end{tikzpicture}
\caption{Canvas rendering pipeline stages.}
\label{fig:render_pipeline}
\end{figure}
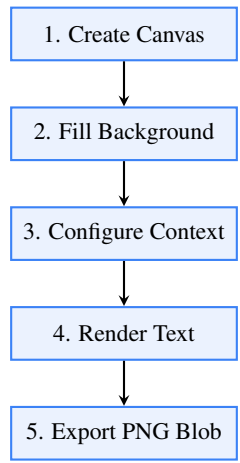

\subsubsection{RTL Text Rendering}

Proper RTL rendering requires canvas context configuration:

\begin{lstlisting}[language=JavaScript, caption={RTL text rendering on canvas.}, label={lst:rtl}]
function renderText(
    ctx: CanvasRenderingContext2D,
    text: string,
    config: RenderConfig
): void {
    const { width, height, font, fontSize, direction } = config;
    
    // Configure RTL context
    ctx.direction = direction;  // 'rtl' or 'ltr'
    ctx.font = `${fontSize}px "${font}"`;
    ctx.fillStyle = config.textColor;
    ctx.textBaseline = 'middle';
    
    // Calculate position
    let x: number;
    if (direction === 'rtl') {
        ctx.textAlign = 'right';
        x = width - config.padding;
    } else {
        ctx.textAlign = 'left';
        x = config.padding;
    }
    const y = height / 2;
    
    // Render
    ctx.fillText(text, x, y);
}
\end{lstlisting}

\subsubsection{Background Composition}

Background selection supports both solid colors and image textures:

\begin{lstlisting}[language=JavaScript, caption={Background rendering with mix mode.}, label={lst:background}]
async function renderBackground(
    ctx: CanvasRenderingContext2D,
    config: BackgroundConfig,
    rng: SeededRandom
): Promise<void> {
    const { width, height } = ctx.canvas;
    
    if (config.mode === 'color') {
        ctx.fillStyle = config.color;
        ctx.fillRect(0, 0, width, height);
    } else if (config.mode === 'image') {
        const img = await loadImage(config.imageUrl);
        ctx.drawImage(img, 0, 0, width, height);
    } else if (config.mode === 'mix') {
        // Select by percentage distribution
        const bg = selectBackground(config.options, rng);
        await renderBackground(ctx, bg, rng);
    }
}
\end{lstlisting}

\begin{figure}[htbp]
  \centering
  \begin{subfigure}{0.45\linewidth}
    \centering
    \includegraphics[width=\linewidth]{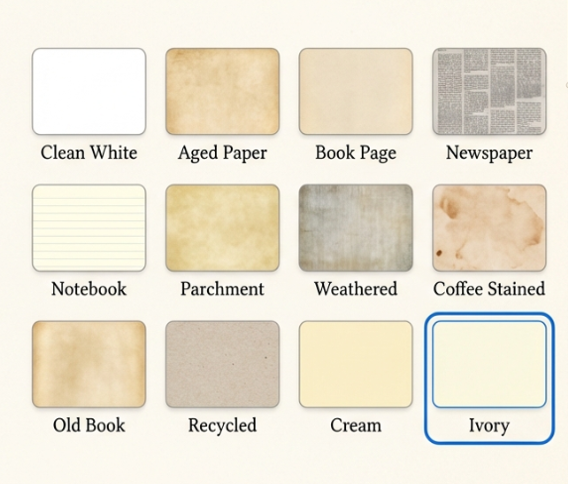}
    \caption{Default color options available for background configuration.}
    \label{fig:background_colors_default}
  \end{subfigure}
  \hfill
  \begin{subfigure}{0.45\linewidth}
    \centering
    \includegraphics[width=\linewidth]{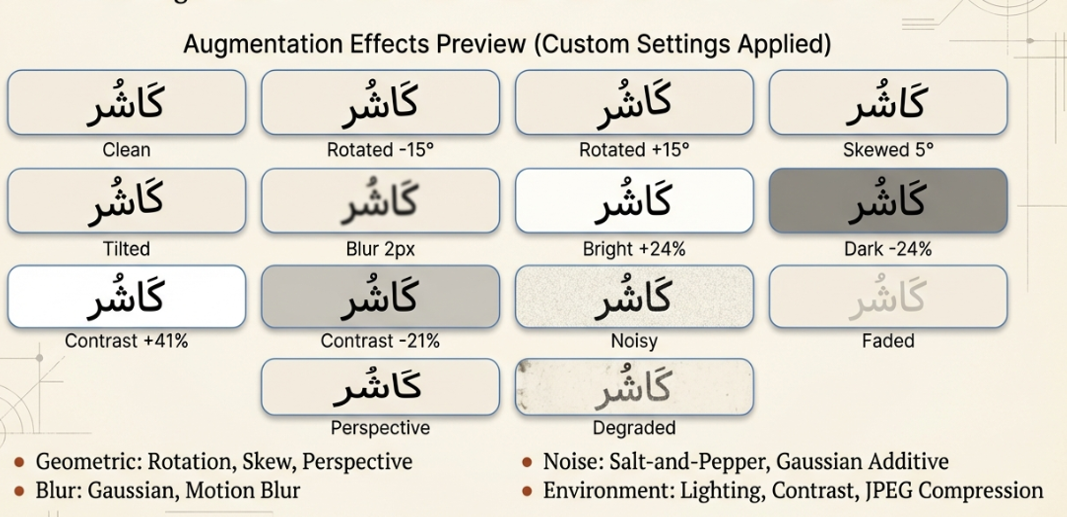}
    \caption{Augmentation settings preview showing available style options.}
    \label{fig:augmentation_settings}
  \end{subfigure}
  \caption{Background configuration interface: (\protect\subref{fig:background_colors_default}) default color palette options, (\protect\subref{fig:augmentation_settings}) augmentation styles settings preview.}
  \label{fig:background_augmentation}
\end{figure}


\subsection{Augmentation Implementation}

\subsubsection{Transform Pipeline Architecture}

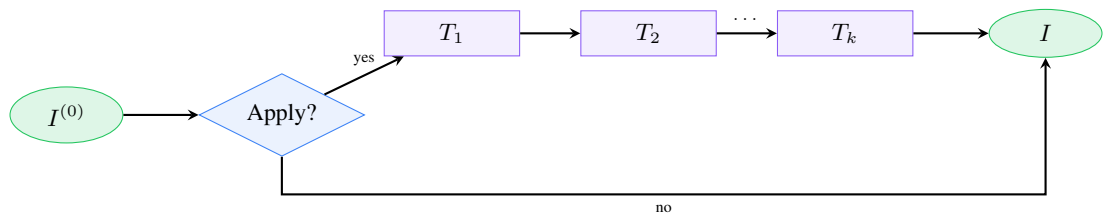
\begin{figure}[htbp]
\centering
\begin{tikzpicture}[
    node distance=0.4cm and 0.8cm,
    input/.style={ellipse, draw=successgreen, fill=successgreen!15,
                  minimum width=1.5cm, align=center, font=\small},
    transform/.style={rectangle, draw=accentpurple, fill=accentpurple!10,
                      minimum width=1.8cm, minimum height=0.6cm, align=center, font=\footnotesize},
    decision/.style={diamond, draw=primaryblue, fill=primaryblue!10,
                     minimum width=1cm, aspect=2, align=center, font=\footnotesize},
    arrow/.style={->, thick, >=stealth}
]

\node[input] (in) {$I^{(0)}$};
\node[decision, right=1cm of in] (d1) {Apply?};
\node[transform, above right=0.5cm and 0.8cm of d1] (t1) {$T_1$};
\node[transform, right=of t1] (t2) {$T_2$};
\node[transform, right=of t2] (t3) {$T_k$};
\node[input, right=1cm of t3] (out) {$I$};

\draw[arrow] (in) -- (d1);
\draw[arrow] (d1) -- node[above, font=\tiny] {yes} (t1);
\draw[arrow] (d1.south) -- ++(0,-0.5) -| node[below, font=\tiny, pos=0.25] {no} (out);
\draw[arrow] (t1) -- (t2);
\draw[arrow] (t2) -- node[above, font=\tiny] {$\cdots$} (t3);
\draw[arrow] (t3) -- (out);

\end{tikzpicture}
\caption{Augmentation transform pipeline with probabilistic application.}
\label{fig:aug_pipeline}
\end{figure}

\subsubsection{Pixel-Level Operations}

Augmentations requiring pixel access use \texttt{ImageData}:

\begin{lstlisting}[language=JavaScript, caption={Gaussian noise injection.}, label={lst:noise}]
function applyGaussianNoise(
    ctx: CanvasRenderingContext2D,
    sigma: number,
    rng: SeededRandom
): void {
    const { width, height } = ctx.canvas;
    const imageData = ctx.getImageData(0, 0, width, height);
    const data = imageData.data;
    
    for (let i = 0; i < data.length; i += 4) {
        // Box-Muller transform for Gaussian samples
        const u1 = rng.next();
        const u2 = rng.next();
        const z = Math.sqrt(-2 * Math.log(u1)) 
                * Math.cos(2 * Math.PI * u2);
        const noise = z * sigma;
        
        // Apply to RGB channels
        data[i]     = clamp(data[i] + noise, 0, 255);
        data[i + 1] = clamp(data[i + 1] + noise, 0, 255);
        data[i + 2] = clamp(data[i + 2] + noise, 0, 255);
    }
    
    ctx.putImageData(imageData, 0, 0);
}

function clamp(val: number, min: number, max: number): number {
    return Math.max(min, Math.min(max, val));
}
\end{lstlisting}

\subsubsection{Geometric Transform Implementation}

Rotation uses canvas context transformation:

\begin{lstlisting}[language=JavaScript, caption={Rotation augmentation.}, label={lst:rotation}]
function applyRotation(
    ctx: CanvasRenderingContext2D,
    angleDegrees: number
): void {
    const { width, height } = ctx.canvas;
    const angleRadians = angleDegrees * Math.PI / 180;
    
    // Translate to center, rotate, translate back
    ctx.translate(width / 2, height / 2);
    ctx.rotate(angleRadians);
    ctx.translate(-width / 2, -height / 2);
}
\end{lstlisting}

The transformation matrix is:
\begin{equation}
M = T_{-c} \cdot R_\theta \cdot T_c = 
\begin{pmatrix}
\cos\theta & -\sin\theta & c_x(1-\cos\theta) + c_y\sin\theta \\
\sin\theta & \cos\theta & c_y(1-\cos\theta) - c_x\sin\theta \\
0 & 0 & 1
\end{pmatrix}
\end{equation}
where $(c_x, c_y) = (W/2, H/2)$ is the image center.

\subsection{Output Generation}

\subsubsection{ZIP Archive Creation}

Archives are created client-side using JSZip:

\begin{lstlisting}[language=JavaScript, caption={ZIP archive generation.}, label={lst:zip}]
async function createArchive(
    samples: Sample[],
    format: OutputFormat
): Promise<Blob> {
    const zip = new JSZip();
    const imagesFolder = zip.folder('images');
    
    // Add images
    for (let i = 0; i < samples.length; i++) {
        const filename = `image_${i.toString().padStart(6, '0')}.png`;
        imagesFolder.file(filename, samples[i].blob);
    }
    
    // Add labels based on format
    const labels = generateLabels(samples, format);
    zip.file(getLabelsFilename(format), labels);
    
    // Add metadata
    zip.file('metadata.json', JSON.stringify(metadata, null, 2));
    
    return zip.generateAsync({
        type: 'blob',
        compression: 'DEFLATE',
        compressionOptions: { level: 6 }
    });
}
\end{lstlisting}

\subsubsection{Label Format Generation}

\begin{lstlisting}[language=JavaScript, caption={Multi-format label generation.}, label={lst:labels}]
function generateLabels(
    samples: Sample[], 
    format: OutputFormat
): string {
    switch (format) {
        case 'crnn':
            return samples.map((s, i) => 
                `image_${i.toString().padStart(6, '0')}.png\t${s.text}`
            ).join('\n');
            
        case 'trocr':
            return samples.map((s, i) => 
                JSON.stringify({
                    image: `images/image_${i.toString().padStart(6, '0')}.png`,
                    text: s.text
                })
            ).join('\n');
            
        case 'huggingface':
            const header = 'file_name,text\n';
            const rows = samples.map((s, i) => 
                `"images/image_${i.toString().padStart(6, '0')}.png","${s.text.replace(/"/g, '""')}"`
            ).join('\n');
            return header + rows;
            
        default:
            throw new Error(`Unknown format: ${format}`);
    }
}
\end{lstlisting}

\subsection{Performance Optimizations}

\subsubsection{Canvas Reuse}

A single canvas element is reused across samples:

\begin{equation}
\text{Memory} = O(H \times W) \text{ constant, rather than } O(N \times H \times W)
\end{equation}

\subsubsection{Incremental ZIP Building}

Images are added incrementally to avoid memory spikes:

\begin{lstlisting}[language=JavaScript, caption={Incremental archive building.}, label={lst:incremental}]
async function generateDataset(
    config: Config,
    onProgress: (p: number) => void
): Promise<Blob> {
    const zip = new JSZip();
    const folder = zip.folder('images');
    
    for (let i = 0; i < config.size; i++) {
        const sample = await generateSample(i, config);
        folder.file(sample.filename, sample.blob);
        
        // Report progress periodically
        if (i % 100 === 0) {
            onProgress(i / config.size);
        }
    }
    
    return zip.generateAsync({ type: 'blob' });
}
\end{lstlisting}

\subsubsection{Complexity Summary}

\begin{table}[htbp]
\centering
\caption{Time and space complexity by operation.}
\label{tab:complexity}
\begin{tabular}{@{}lll@{}}
\toprule
\textbf{Operation} & \textbf{Time} & \textbf{Space} \\
\midrule
Text segmentation & $O(n)$ & $O(M \cdot \bar{\ell})$ \\
Unicode validation & $O(M \cdot \bar{\ell})$ & $O(1)$ \\
Single image render & $O(H \cdot W)$ & $O(H \cdot W)$ \\
Augmentation (per image) & $O(H \cdot W \cdot k)$ & $O(H \cdot W)$ \\
Label generation & $O(N \cdot \bar{\ell})$ & $O(N \cdot \bar{\ell})$ \\
ZIP compression & $O(N \cdot H \cdot W)$ & $O(N \cdot H \cdot W / r)$ \\
\midrule
\textbf{Total} & $O(N \cdot H \cdot W \cdot k)$ & $O(N \cdot H \cdot W / r)$ \\
\bottomrule
\end{tabular}
\end{table}

where $n$ = corpus size, $M$ = segment count, $\bar{\ell}$ = average segment length, $N$ = dataset size, $H \times W$ = image dimensions, $k$ = max augmentations, $r$ = compression ratio.

\section{Memory Management and Sample Storage}
\label{sec:Memory_Management_and_Sample_Storage}

This section details how SynthOCR-Gen manages memory during generation and the various sample storage strategies available.
As a browser-based application, SynthOCR-Gen operates within browser memory constraints:

\begin{table}[htbp]
\centering
\caption{Browser memory characteristics.}
\label{tab:browser_memory}
\begin{tabular}{@{}ll@{}}
\toprule
\textbf{Characteristic} & \textbf{Typical Value} \\
\midrule
JavaScript heap limit (Chrome) & 2--4 GB \\
JavaScript heap limit (Firefox) & 2--8 GB \\
Single ArrayBuffer limit & 2 GB \\
Canvas max dimensions & 32,767 × 32,767 px \\
Recommended max dataset & 100K--500K samples \\
\bottomrule
\end{tabular}
\end{table}

\subsubsection{Memory Usage Per Sample}

For a single sample at $256 \times 64$ pixels (typical OCR dimensions):

\begin{align}
\text{Canvas pixels} &= 256 \times 64 = 16,384 \text{ pixels} \\
\text{Raw RGBA data} &= 16,384 \times 4 \text{ bytes} = 65,536 \text{ bytes} \approx 64 \text{ KB} \\
\text{PNG compressed} &\approx 5\text{--}15 \text{ KB per image} \\
\text{Label string} &\approx 20\text{--}100 \text{ bytes}
\end{align}

\subsubsection{Storage Modes}

SynthOCR-Gen supports three sample storage strategies during generation:

\begin{table}[htbp]
\centering
\caption{Sample storage modes comparison.}
\label{tab:storage_modes}
\begin{tabular}{@{}lp{4cm}p{4cm}l@{}}
\toprule
\textbf{Mode} & \textbf{Description} & \textbf{Best For} & \textbf{Memory} \\
\midrule
\textbf{In-Memory ZIP} & All samples accumulated in JSZip object before download & Small to medium datasets (< 100K) & $O(N)$ \\
\midrule
\textbf{Streaming/Chunked} & Samples processed in batches, partial ZIPs created & Large datasets (100K--500K) & $O(\text{batch})$ \\
\midrule
\textbf{Individual Files} & Each sample saved separately (CLI mode) & Very large datasets (> 500K) & $O(1)$ \\
\bottomrule
\end{tabular}
\end{table}

\paragraph{Mode 1: In-Memory ZIP (Default).}

The default mode accumulates all samples in memory before creating the final ZIP archive:

\begin{lstlisting}[language=JavaScript, caption={In-memory ZIP accumulation.}, label={lst:inmemory}]
// All samples stored in memory
const zip = new JSZip();
const folder = zip.folder('images');

for (let i = 0; i < N; i++) {
    const sample = await generateSample(i);
    folder.file(sample.filename, sample.blob);  // Blob held in memory
}

// Only releases memory after download
const archive = await zip.generateAsync({ type: 'blob' });
saveAs(archive, 'dataset.zip');
\end{lstlisting}

\textbf{Memory usage:} $O(N \times \text{avg\_image\_size})$

\textbf{Best for:} Datasets under 100,000 samples on machines with 8+ GB RAM.

\paragraph{Mode 2: Streaming/Chunked Generation.}

For larger datasets, samples are processed in batches with periodic cleanup:

\begin{lstlisting}[language=JavaScript, caption={Chunked generation with memory cleanup.}, label={lst:chunked}]
const BATCH_SIZE = 10000;

async function generateInBatches(
    config: Config
): Promise<Blob[]> {
    const batches: Blob[] = [];
    
    for (let batch = 0; batch < Math.ceil(config.size / BATCH_SIZE); batch++) {
        const zip = new JSZip();
        const folder = zip.folder('images');
        
        const start = batch * BATCH_SIZE;
        const end = Math.min(start + BATCH_SIZE, config.size);
        
        for (let i = start; i < end; i++) {
            const sample = await generateSample(i);
            folder.file(sample.filename, sample.blob);
        }
        
        // Create partial ZIP and release memory
        const batchZip = await zip.generateAsync({ type: 'blob' });
        batches.push(batchZip);
        
        // Allow garbage collection
        await new Promise(r => setTimeout(r, 100));
    }
    
    return batches;  // Multiple ZIP files
}
\end{lstlisting}

\textbf{Memory usage:} $O(\text{BATCH\_SIZE} \times \text{avg\_image\_size})$

\textbf{Best for:} Datasets of 100,000--500,000 samples.

\paragraph{Mode 3: Individual File Saving (CLI Mode).}

The CLI version writes each sample directly to disk:

\begin{lstlisting}[language=JavaScript, caption={Direct file system writing (Node.js CLI).}, label={lst:filesystem}]
import { writeFile, mkdir } from 'fs/promises';
import sharp from 'sharp';

async function generateToFileSystem(
    config: Config,
    outputDir: string
): Promise<void> {
    await mkdir(`${outputDir}/images`, { recursive: true });
    const labels: string[] = [];
    
    for (let i = 0; i < config.size; i++) {
        const { imageBuffer, text } = await generateSample(i);
        
        const filename = `image_${i.toString().padStart(6, '0')}.png`;
        
        // Write directly to disk - no memory accumulation
        await writeFile(`${outputDir}/images/${filename}`, imageBuffer);
        labels.push(`${filename}\t${text}`);
        
        // Optionally flush labels periodically
        if (i % 10000 === 0) {
            await writeFile(
                `${outputDir}/labels_partial.txt`, 
                labels.join('\n')
            );
        }
    }
    
    await writeFile(`${outputDir}/labels.txt`, labels.join('\n'));
}
\end{lstlisting}

\textbf{Memory usage:} $O(1)$ constant --- only one sample in memory at a time.

\textbf{Best for:} Very large datasets (> 500,000 samples) or memory-constrained environments.

\subsubsection{Memory Optimization Techniques}

SynthOCR-Gen employs several techniques to minimize memory usage:

\begin{enumerate}
    \item \textbf{Canvas Reuse}: A single canvas element is cleared and reused for each sample rather than creating new DOM elements.
    
    \item \textbf{Blob Conversion}: Image data is immediately converted to compressed PNG Blob format, reducing memory footprint by 4--10x compared to raw RGBA.
    
    \item \textbf{Lazy Font Loading}: Fonts are loaded on-demand and cached, rather than loading all fonts upfront.
    
    \item \textbf{Progress Callbacks}: The UI can interrupt generation to allow garbage collection cycles.
    
    \item \textbf{Incremental ZIP Building}: Samples are added to the ZIP archive incrementally rather than building a large array first.
\end{enumerate}

\subsubsection{Storage Format Comparison}

\begin{table}[htbp]
\centering
\caption{Output format storage characteristics.}
\label{tab:format_storage}
\begin{tabular}{@{}lrrr@{}}
\toprule
\textbf{Format} & \textbf{Avg Image Size} & \textbf{Label Overhead} & \textbf{100K Dataset} \\
\midrule
PNG (default) & 8--12 KB & --- & 0.8--1.2 GB \\
JPEG (quality 85) & 3--6 KB & --- & 0.3--0.6 GB \\
WebP (quality 80) & 2--4 KB & --- & 0.2--0.4 GB \\
\midrule
CRNN (labels.txt) & --- & ~30 bytes/sample & ~3 MB \\
TrOCR (data.jsonl) & --- & ~80 bytes/sample & ~8 MB \\
HuggingFace (metadata.csv) & --- & ~60 bytes/sample & ~6 MB \\
\bottomrule
\end{tabular}
\end{table}

\subsubsection{Recommendations by Dataset Size}

\begin{table}[htbp]
\centering
\caption{Recommended configuration by dataset size.}
\label{tab:size_recommendations}
\begin{tabular}{@{}llll@{}}
\toprule
\textbf{Dataset Size} & \textbf{Storage Mode} & \textbf{Min RAM} & \textbf{Est. Time} \\
\midrule
< 10,000 & In-Memory ZIP & 4 GB & < 5 min \\
10K -- 50K & In-Memory ZIP & 8 GB & 5--20 min \\
50K -- 100K & In-Memory ZIP & 16 GB & 20--45 min \\
100K -- 250K & Chunked (10K batches) & 8 GB & 45--120 min \\
250K -- 500K & Chunked (10K batches) & 8 GB & 2--4 hours \\
500K -- 1M & CLI + Filesystem & 4 GB & 4--8 hours \\
> 1M & CLI + Filesystem & 4 GB & 8+ hours \\
\bottomrule
\end{tabular}
\end{table}

\subsubsection{Browser Memory Monitoring}

During generation, the tool monitors memory usage via the Performance API:

\begin{lstlisting}[language=JavaScript, caption={Memory monitoring and automatic batching.}, label={lst:memmonitor}]
function checkMemoryPressure(): boolean {
    if ('memory' in performance) {
        const memory = (performance as any).memory;
        const usedRatio = memory.usedJSHeapSize / memory.jsHeapSizeLimit;
        
        // Trigger batching if > 70% memory used
        return usedRatio > 0.7;
    }
    return false;  // API not available
}

async function generateWithMemoryGuard(config: Config): Promise<Blob> {
    const zip = new JSZip();
    
    for (let i = 0; i < config.size; i++) {
        if (checkMemoryPressure()) {
            // Force garbage collection opportunity
            await new Promise(r => setTimeout(r, 500));
            console.warn(`Memory pressure at sample ${i}, pausing...`);
        }
        
        const sample = await generateSample(i);
        zip.folder('images').file(sample.filename, sample.blob);
    }
    
    return zip.generateAsync({ type: 'blob' });
}
\end{lstlisting}

\subsubsection{Sample Integrity Verification}

All saved samples include integrity checks:

\begin{itemize}[noitemsep]
    \item \textbf{Image validation}: PNG headers verified before adding to archive
    \item \textbf{Label consistency}: Sample count matches label line count
    \item \textbf{Unicode preservation}: Ground-truth labels use UTF-8 encoding with BOM for compatibility
    \item \textbf{Metadata logging}: Generation parameters, timestamps, and checksums stored in \texttt{metadata.json}
\end{itemize}

\section{Experiments and Results}
\label{sec:experiments}

We validate SynthOCR-Gen by generating a large-scale Kashmiri OCR dataset and analyzing its characteristics. While Kashmiri serves as our case study due to its status as an underserved low-resource language with no existing OCR support, the methodology demonstrated here is entirely language-agnostic and applicable to any Unicode-supported writing system.

\begin{tcolorbox}[colback=blue!5!white, colframe=blue!75!black, title=\textbf{Note on Language Universality}]
\textbf{Important:} While we use Kashmiri as our case study, SYNTHOCR-GEN: is \textbf{language-agnostic}. The tool accepts \textit{any} Unicode-encoded text as input. Kashmiri was chosen for demonstration and validation purposes due to its status as an underserved low-resource language. Researchers working with other scripts including Devanagari, Tibetan, Ethiopic, Georgian, Armenian, or any other Unicode-supported writing system can apply the same methodology by simply providing their own digital Unicode text corpus and appropriate fonts.
\end{tcolorbox}
\subsection{Experimental Setup}

Our source text was sampled from the KS-LIT-3M dataset \citep{kslit3m2025}, a 3.1 million word Kashmiri text corpus designed for large language model pretraining. To our knowledge, KS-LIT-3M represents the largest publicly available Kashmiri text dataset, comprising digitized literary works, journalistic content, and contemporary writing in Unicode-encoded Kashmiri script. For dataset generation, we extracted a representative sample and applied preprocessing including Unicode normalization to NFC form, removal of control characters and formatting artifacts, and script purity verification to ensure 100\% Persio-Arabic-script content. The corpus includes prose, poetry, and journalistic text, providing stylistic diversity representative of contemporary Kashmiri writing.

We selected Kashmiri as our case study for several compelling reasons. First, it represents a genuinely low-resource language with no existing OCR support in Tesseract, TrOCR, or PaddleOCR, making it an ideal test case for our synthetic data approach. Second, its complex Perso-Arabic script with unique diacritics and extended characters presents meaningful technical challenges that validate our Unicode handling capabilities. Third, the availability of KS-LIT-3M provides sufficient high-quality source text for generation, and the published nature of this dataset enables other researchers to replicate our experiments. However, we emphasize that the methodology is entirely transferable to any language with Unicode text availability and appropriate fonts.

Table~\ref{tab:corpus_config} presents the source corpus statistics and generation configuration. Generation was performed on a consumer-grade machine equipped with an Intel Core i7-12700H processor, 16 GB DDR4 RAM, and Chrome 120 running on Windows 11, demonstrating that our browser-based approach does not require specialized hardware. Total generation time for 600,000 samples was approximately 4.5 hours, yielding an average rate of 37 samples per second, with the resulting ZIP archive size of 8.7 GB uncompressed.

\begin{table}[htbp]
\centering
\caption{Source corpus statistics and generation configuration.}
\label{tab:corpus_config}
\begin{tabular}{@{}llll@{}}
\toprule
\multicolumn{2}{c}{\textbf{Source Corpus}} & \multicolumn{2}{c}{\textbf{Generation Parameters}} \\
\midrule
Source dataset & KS-LIT-3M \citep{kslit3m2025} & Dataset size & 600,000 samples \\
Sample size & 487,216 words & Segmentation mode & Word-level \\
Unique words & 89,743 & Image dimensions & 256 × 64 pixels \\
Total characters & 2,847,329 & Random seed & 42 \\
Total sentences & 31,562 & Train/Val split & 90\% / 10\% \\
File size & 3.2 MB (UTF-8 NFC) & Augmentation rate & 70\% \\
\bottomrule
\end{tabular}
\end{table}

\subsection{Dataset Characteristics}

The generated dataset comprises 600,000 image-text pairs divided into a training set of 540,000 samples (90\%) and a validation set of 60,000 samples (10\%). Of the total samples, 180,000 (30\%) are clean renderings without augmentation, while the remaining 420,000 (70\%) include various augmentation effects. The text content exhibits natural variation with an average of 6.8 characters per sample, a median of 5 characters, and a range from 1 to 24 characters. Notably, 523,412 samples (87.2\%) contain diacritical marks, reflecting the importance of diacritics in Kashmiri orthography and ensuring that trained models learn to recognize these critical features. This dataset is published separately at \url{https://arxiv.org/abs/2601.01088}.

Font distribution across the generated samples closely matches the configured percentages, with Noto Naskh Arabic appearing in 40.1\% of samples, Gulmarg Nastaleeq in 34.8\%, and Scheherazade New in 25.1\%. This multi-font approach ensures that trained models generalize across different typographic styles commonly encountered in Kashmiri printed materials. Background textures follow the configured distribution with Clean White at 30\%, Aged Paper at 25\%, Book Page at 20\%, Newspaper at 15\%, and Parchment at 10\%, simulating the variety of document conditions encountered in real-world OCR applications.

Table~\ref{tab:aug_char} presents the augmentation application statistics and character coverage analysis. Among augmented samples, rotation was the most frequently applied transform (40.1\%), followed by brightness variation (35.1\%), Gaussian blur (30.2\%), and Gaussian noise (26.1\%). On average, each augmented sample received 2.3 transformations out of a maximum of 4, indicating effective probabilistic sampling across the augmentation space. Character coverage analysis confirms that all 85 unique characters in the Kashmiri writing system are represented in the dataset, including 38 base Arabic letters, 8 Kashmiri extended letters, 12 Arabic diacritics, 6 Kashmiri-specific diacritics, 11 punctuation marks, and 10 numerals.

\begin{table}[htbp]
\centering
\caption{Augmentation statistics and character coverage.}
\label{tab:aug_char}
\begin{tabular}{@{}lrr|lrr@{}}
\toprule
\textbf{Augmentation} & \textbf{Count} & \textbf{\%} & \textbf{Character Category} & \textbf{Unique} & \textbf{Occurrences} \\
\midrule
Rotation & 168,420 & 40.1\% & Base Arabic letters & 38 & 3,241,567 \\
Brightness variation & 147,336 & 35.1\% & Kashmiri extended & 8 & 412,893 \\
Gaussian blur & 126,672 & 30.2\% & Arabic diacritics & 12 & 2,187,432 \\
Gaussian noise & 109,620 & 26.1\% & Kashmiri diacritics & 6 & 523,412 \\
Contrast variation & 92,820 & 22.1\% & Punctuation & 11 & 156,234 \\
JPEG artifacts & 63,084 & 15.0\% & Numerals & 10 & 89,234 \\
\midrule
Avg. per sample & 2.3 & --- & \textbf{Total unique} & \textbf{85} & --- \\
\bottomrule
\end{tabular}
\end{table}

\subsection{Sample Visualization}

\begin{figure}[htbp]
\centering
\begin{minipage}[b]{0.48\textwidth}
\centering
\includegraphics[width=\textwidth]{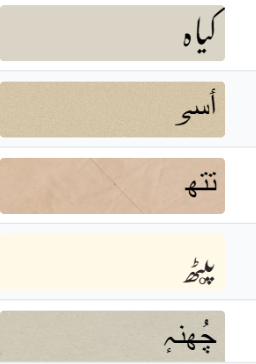}
\caption{Rendering pipeline output visualization. The final image is produced by compositing the selected font rendering over the background texture, demonstrating the text-to-image generation process that produces perfectly-labeled training data.}
\label{fig:samples}
\end{minipage}
\hfill
\begin{minipage}[b]{0.48\textwidth}
\centering
\includegraphics[width=\textwidth]{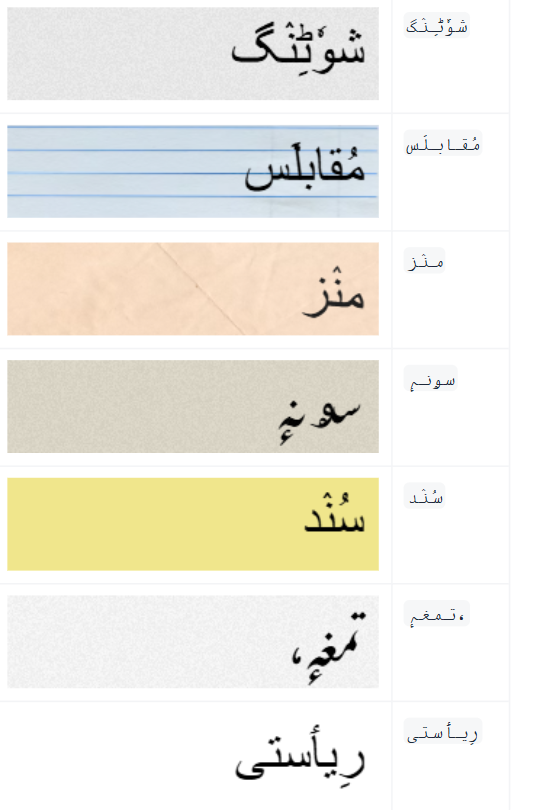}
\caption{Sample images from the 600K Kashmiri OCR dataset showing generated word images with their corresponding ground-truth labels. The samples demonstrate variation in fonts (Noto Naskh, Gulmarg Nastalik, Scheherazade), backgrounds (clean, aged paper, book page), and slight augmentation effects (rotation,noise)}
\label{fig:rendering_output}
\end{minipage}

\end{figure}

\subsection{Deployment and Reproducibility}

Both the generated dataset and the SynthOCR-Gen tool have been publicly deployed to facilitate community adoption and reproducibility. The 600K Kashmiri OCR dataset is available on HuggingFace Hub at \url{https://huggingface.co/datasets/Omarrran/600k_KS_OCR_Word_Segmented_Dataset}, where it can be loaded directly using the HuggingFace \texttt{datasets} library . The dataset viewer on HuggingFace Hub provides browsable sample visualization with images and corresponding ground-truth transcriptions for easy inspection.

The SynthOCR-Gen tool is deployed on HuggingFace Spaces at \url{https://huggingface.co/spaces/Omarrran/OCR_DATASET_MAKER}, enabling researchers to generate custom datasets for any Unicode-supported language without local installation. Users simply upload their Unicode text file and compatible fonts for their script, configure generation parameters according to their requirements, and download the generated dataset ready for OCR model training.

To verify reproducibility, we regenerated the dataset with identical configuration and seed. Comparison of the two archives confirmed identical file names and counts, identical label file contents with byte-for-byte matching, and identical image pixel values as verified via hash comparison, confirming that seeded randomization achieves complete reproducibility. Table~\ref{tab:performance} presents generation performance benchmarks across different dataset sizes, demonstrating consistent throughput until very large datasets where browser memory pressure causes slight slowdown.

\begin{table}[htbp]
\centering
\caption{Generation performance benchmarks across dataset sizes.}
\label{tab:performance}
\begin{tabular}{@{}lrrr@{}}
\toprule
\textbf{Dataset Size} & \textbf{Time} & \textbf{Rate (samples/sec)} & \textbf{ZIP Size} \\
\midrule
10,000 & 4 min & 42 & 145 MB \\
50,000 & 19 min & 44 & 725 MB \\
100,000 & 38 min & 44 & 1.45 GB \\
250,000 & 96 min & 43 & 3.6 GB \\
600,000 & 270 min & 37 & 8.7 GB \\
\bottomrule
\end{tabular}
\end{table}

\subsection{Applicability to Other Languages}

While our experiments focus on Kashmiri, the SynthOCR-Gen methodology is designed for broad applicability across the world's writing systems. Table~\ref{tab:other_languages} lists example languages that could benefit from similar treatment, representing a combined speaker population exceeding 250 million people currently underserved by OCR technology. Any language with an available Unicode text corpus and appropriate fonts can leverage our tool for synthetic OCR dataset generation, requiring only configuration of text direction (RTL or LTR) and specification of valid Unicode ranges for script purity validation.

\begin{table}[htbp]
\centering
\caption{Example languages compatible with SynthOCR-Gen methodology.}
\label{tab:other_languages}
\begin{tabular}{@{}llll@{}}
\toprule
\textbf{Language} & \textbf{Script} & \textbf{Unicode Block} & \textbf{Direction} \\
\midrule
Kashmiri (this work) & Perso-Arabic & Arabic (0600--06FF) & RTL \\
Urdu & Perso-Arabic & Arabic (0600--06FF) & RTL \\
Pashto & Perso-Arabic & Arabic (0600--06FF) & RTL \\
Hindi & Devanagari & Devanagari (0900--097F) & LTR \\
Tibetan & Tibetan & Tibetan (0F00--0FFF) & LTR \\
Amharic & Ethiopic & Ethiopic (1200--137F) & LTR \\
Georgian & Georgian & Georgian (10A0--10FF) & LTR \\
Armenian & Armenian & Armenian (0530--058F) & LTR \\
\bottomrule
\end{tabular}
\end{table}

\section{Discussion}
\label{sec:discussion}

This section examines the implications of our work, analyzes the trade-offs between synthetic and real data approaches, discusses the scalability of our methodology to other languages, and acknowledges the limitations of the current implementation.

\subsection{Advantages of Synthetic Data Generation}

The economics of OCR dataset creation are fundamentally transformed by synthetic generation. As illustrated in Table~\ref{tab:cost_comparison}, manual annotation of a 600,000-sample dataset would require an estimated 20+ person-months of labor and cost between \$50,000 and \$100,000, assuming professional annotators working with complex Arabic-script text. The process would span 6--12 months and yield datasets with inherent human error rates of 1--5\%. In contrast, synthetic generation produces the same volume of data in approximately 4.5 hours at essentially zero marginal cost, with mathematically perfect ground-truth labels guaranteed by the generation process itself.

\begin{table}[htbp]
\centering
\caption{Cost comparison: Manual vs. Synthetic dataset creation for 600K samples.}
\label{tab:cost_comparison}
\begin{tabular}{@{}lrr@{}}
\toprule
\textbf{Resource} & \textbf{Manual} & \textbf{Synthetic} \\
\midrule
Human annotators & 20+ person-months & 0 \\
Annotation cost & \$50,000--100,000 & \$0 \\
Quality assurance & Extensive & N/A \\
Generation time & 6--12 months & 4.5 hours \\
Reproducibility & Low & Perfect \\
Scalability & Linear cost & Near-zero marginal cost \\
\bottomrule
\end{tabular}
\end{table}

Beyond cost efficiency, synthetic generation provides unprecedented control over dataset characteristics. Researchers can precisely tune font distributions to match expected deployment conditions, calibrate augmentation severity according to training curriculum needs, balance character and word frequencies to address class imbalance, and deliberately oversample rare characters that would be underrepresented in natural text. This level of control is simply impossible with real-world data collection, where the distribution of samples is determined by whatever documents happen to be available.

The reproducibility advantage deserves particular emphasis. With seeded randomization, any researcher can regenerate an identical dataset given the same inputs and configuration. This enables rigorous experimental comparisons, debugging of training issues, and long-term archival of experimental conditions, capabilities that are difficult or impossible to achieve with manually curated datasets.

\subsection{Synthetic vs. Real Data Trade-offs}

The primary concern with synthetic data is the potential domain gap between training images and real-world test images. While our augmentation pipeline simulates many common degradations, including rotation, blur, noise, compression artifacts, and uneven lighting, it cannot capture all variations present in authentic scanned documents, photographs, or handwritten text. Physical paper creases, fold marks, water damage, non-uniform scanning artifacts from aged equipment, and the organic variability of handwriting remain challenging to synthesize convincingly.

However, prior work on English OCR has demonstrated that models pre-trained on synthetic data and fine-tuned on small real datasets often outperform models trained exclusively on either data type \citep{souibgui2020synthetic}. This suggests a practical workflow: use synthetic data for initial model development and capability building, then refine with limited real-world samples for deployment. Several strategies can further mitigate domain gap effects, including aggressive augmentation during training to improve generalization, domain adaptation through fine-tuning on target-domain samples, mixed training that combines synthetic and real data, and test-time augmentation to align inference conditions with training distributions.

For low-resource languages where no existing OCR capability exists, the domain gap concern is secondary to the fundamental problem of having any training data at all. Synthetic data provides a starting point from which incremental improvements can be made as real-world samples become available.

\subsection{Scalability to Other Languages}

While we have focused on Kashmiri as our case study, the SynthOCR-Gen methodology is designed for broad applicability. Extending the approach to a new language requires only four components: a Unicode text corpus in any format (books, articles, web scrapes), appropriate TrueType or OpenType fonts that support the target script, configuration of text direction (RTL or LTR) if applicable, and specification of valid Unicode ranges for script purity validation.

We have validated the system with several additional scripts beyond Kashmiri, including Modern Standard Arabic, Persian/Farsi, Urdu, and Hindi in Devanagari script. The modular architecture supports any Unicode-encodable writing system for which fonts are available. Table~\ref{tab:target_languages} lists additional languages that could benefit from similar treatment, representing a combined speaker population of over 250 million people currently underserved by OCR technology.

\begin{table}[htbp]
\centering
\caption{Potential target languages for synthetic OCR dataset generation.}
\label{tab:target_languages}
\begin{tabular}{@{}lll@{}}
\toprule
\textbf{Language} & \textbf{Script} & \textbf{Speakers} \\
\midrule
Pashto & Arabic-based & 50M \\
Sindhi & Arabic-based & 30M \\
Kurdish & Arabic-based & 30M \\
Uyghur & Arabic-based & 15M \\
Balochi & Arabic-based & 8M \\
Punjabi (Shahmukhi) & Arabic-based & 80M \\
Dzongkha & Tibetan & 0.6M \\
Tibetan & Tibetan & 6M \\
Khmer & Khmer & 16M \\
Burmese & Myanmar & 35M \\
\bottomrule
\end{tabular}
\end{table}

\subsection{Limitations}

Several limitations of the current implementation warrant acknowledgment. First, our system focuses exclusively on printed text with standardized digital fonts; handwriting recognition, which requires modeling individual writing style variations, lies outside the current scope. Extending to handwriting would require either handwriting-style fonts (which offer limited diversity), neural handwriting synthesis models (which are computationally expensive), or trajectory-based stroke simulation approaches.

Second, while we support multiple segmentation modes, our Kashmiri dataset uses word-level segmentation. This granularity suits most modern OCR architectures based on CTC or attention mechanisms, but may not optimally serve character-level recognition training, full-page document layout analysis, or structured document understanding tasks involving tables and forms.

Third, the quality of synthetic data depends heavily on font availability. For extremely rare or historical scripts, high-quality digital fonts may be scarce or nonexistent. Open-source font projects such as Google Noto and SIL International have dramatically improved coverage, but gaps remain for archaic scripts and minority writing systems.

Finally, the client-side browser architecture imposes practical memory limits. Very large text corpora exceeding 100MB may challenge browser memory constraints, and datasets approaching one million samples may require generation in batches. For such scales, the CLI mode with direct filesystem access provides an alternative.

\subsection{Privacy and Ethical Considerations}

The fully client-side architecture provides strong privacy guarantees: user text corpora are processed entirely within the browser and never transmitted to external servers. This design is particularly important for applications involving culturally sensitive religious texts, personal correspondence digitization, and government document processing where data sovereignty is paramount.

We acknowledge that OCR technology, including the training data that enables it, could potentially be misused for surveillance purposes such as automated analysis of private correspondence. However, we believe the benefits to language preservation, accessibility for visually impaired users, and cultural heritage digitization substantially outweigh these risks. The open-source nature of our tool ensures transparency about its capabilities and limitations.

For endangered and minority languages, OCR capability supports crucial preservation activities including digitization of historical documents and manuscripts, development of accessibility tools such as screen readers and document-to-speech systems, creation of educational technology for language learning, and archival efforts that protect linguistic heritage against loss. These applications align with broader goals of linguistic diversity preservation and digital inclusion.

\subsection{Future Directions}

Several promising directions for future research emerge from this work. Integration of neural handwriting synthesis models, particularly recent advances in diffusion-based text generation, could extend the approach to manuscript digitization and historical document processing. Development of full-page document layout generation would enable training complete OCR pipelines that include text detection, layout analysis, and recognition stages. Support for multi-script generation would serve language communities where multiple writing systems coexist, such as Kashmiri written in both Perso-Arabic and Devanagari scripts. Finally, integration with active learning frameworks could optimize the balance between synthetic pre-training and targeted real-world annotation, identifying samples where model uncertainty is highest and annotation effort would be most valuable.

\section{Conclusion}
\label{sec:conclusion}

This paper has presented SynthOCR-Gen, an open-source synthetic OCR dataset generator designed to address the critical data scarcity problem that has long impeded OCR development for low-resource languages. By enabling the generation of large-scale, perfectly-labeled training datasets from existing Unicode text corpora, our work removes one of the primary barriers to OCR capability for the world's underserved writing systems.

\subsection{Summary of Contributions}

Our work makes four principal contributions to the field of optical character recognition and low-resource language processing.

First, we developed SynthOCR-Gen, a comprehensive web-based application for generating synthetic OCR training datasets. The tool operates entirely client-side within modern web browsers, ensuring user privacy while eliminating installation barriers. It supports five text segmentation modes ranging from single characters to complete sentences, enabling generation of training data appropriate for various OCR architectures. The system handles multiple fonts with configurable percentage-based distribution, implements over 25 data augmentation techniques that simulate real-world document degradations, provides native support for right-to-left scripts with proper Arabic diacritic preservation, and outputs datasets in formats compatible with major OCR frameworks including CRNN, TrOCR, PaddleOCR, and HuggingFace Transformers. Seeded randomization ensures that any generated dataset can be perfectly reproduced given identical inputs.

Second, we generated and publicly released a 600,000-sample word-segmented Kashmiri OCR dataset, representing the first large-scale OCR dataset for this language. The dataset includes samples rendered in three Arabic-script fonts with diverse background textures and realistic augmentations. It is available on HuggingFace Hub for immediate use by the research community, providing a foundation for developing OCR systems that could serve the 7 million speakers of Kashmiri worldwide.

Third, we established a replicable methodology for creating synthetic OCR datasets applicable to any low-resource language with available Unicode text and appropriate fonts. The approach has been validated on multiple scripts and can be immediately applied to hundreds of languages currently lacking OCR support.

Fourth, we addressed specific technical challenges that arise in Arabic-script text processing, including Kashmiri-specific Unicode diacritic handling, proper grapheme cluster segmentation using the Intl.Segmenter API, and correct right-to-left canvas rendering in browser environments. These solutions are documented and available for reuse.

\subsection{Impact and Significance}

The immediate impact of this work is the enablement of Kashmiri OCR development that was previously blocked by the complete unavailability of training data. More broadly, SynthOCR-Gen reduces the barrier to entry for OCR research on low-resource languages from months of expensive annotation effort to hours of computation on consumer hardware. Research groups and community organizations can now create substantial OCR datasets without external funding or specialized infrastructure.

The methodology demonstrated here provides a template for community-driven dataset creation. Language communities themselves can generate OCR training data using digitized texts in their own languages, fonts familiar from their own printed materials, and configurations tailored to their specific document processing needs. This democratization of dataset creation has the potential to accelerate OCR development across dozens of currently underserved languages.

\subsection{Future Research Directions}

Several promising directions for future research emerge from this work. Extension to handwritten text synthesis, potentially leveraging recent advances in diffusion-based generative models, would enable manuscript digitization and historical document processing applications. Development of full-page document layout generation capabilities would support training of complete OCR pipelines including text detection and layout analysis stages. Multi-script support for languages written in multiple writing systems would serve communities with script plurality. Integration with active learning frameworks could optimize the balance between synthetic pre-training and targeted real-world annotation. Finally, systematic evaluation of models trained on our synthetic data against real Kashmiri documents would quantify transfer learning effectiveness and guide future improvements.

\subsection{Resource Availability}

All resources developed in this work are publicly available to support reproducibility and community adoption. The SynthOCR-Gen tool is deployed at \url{https://huggingface.co/spaces/Omarrran/OCR_DATASET_MAKER}. The 600K Kashmiri OCR dataset is available at \url{https://huggingface.co/datasets/Omarrran/600k_KS_OCR_Word_Segmented_Dataset}. The complete source code is released under the MIT license at \url{https://github.com/HAQ-NAWAZ-MALIK/OCR_TEXT_RECOG_DATASET_MAKER}.

\subsection{Closing Remarks}

The digital marginalization of low-resource languages need not persist. While the gap between high-resource and low-resource language AI capabilities remains substantial, tools like SynthOCR-Gen demonstrate that practical paths forward exist. Synthetic data generation offers a bridge, not a replacement for real-world data, but a means of enabling initial capability development that can subsequently be refined through targeted annotation.

We hope that SynthOCR-Gen and the Kashmiri dataset serve as useful resources for the research community and as templates for similar efforts across the world's diverse writing systems. OCR capability is a fundamental building block for digital accessibility, archival preservation, and AI integration. By democratizing OCR dataset creation, we contribute toward the goal of universal text recognition capability for all of the world's languages.

\section*{Acknowledgments}
We thank the open-source community for their contributions to the tools and libraries used in this work. We also thank the Kashmiri language community for their efforts in preserving and digitizing their linguistic heritage.

\bibliographystyle{unsrtnat}
\bibliography{references}

@inproceedings{jaderberg2014synthetic,
  title={Synthetic Data and Artificial Neural Networks for Natural Scene Text Recognition},
  author={Jaderberg, Max and Simonyan, Karen and Vedaldi, Andrea and Zisserman, Andrew},
  booktitle={Advances in Neural Information Processing Systems (NIPS) Workshop on Deep Learning},
  year={2014},
  note={Introduced MJSynth dataset with 8.9M synthetic word images}
}

@inproceedings{gupta2016synthetic,
  title={Synthetic Data for Text Localisation in Natural Images},
  author={Gupta, Ankush and Vedaldi, Andrea and Zisserman, Andrew},
  booktitle={Proceedings of the IEEE Conference on Computer Vision and Pattern Recognition (CVPR)},
  pages={2315--2324},
  year={2016},
  note={Introduced SynthText dataset for scene text detection}
}

@inproceedings{yim2021synthtiger,
  title={SynthTIGER: Synthetic Text Image GEneratoR Towards Better Text Recognition Models},
  author={Yim, Moonbin and Kim, Yoonsik and Cho, Han-Cheol and Park, Sungrae},
  booktitle={Proceedings of the International Conference on Document Analysis and Recognition (ICDAR)},
  pages={109--124},
  year={2021},
  organization={Springer}
}

@article{li2021trocr,
  title={TrOCR: Transformer-based Optical Character Recognition with Pre-trained Models},
  author={Li, Minghao and Lv, Tengchao and Chen, Jingye and Cui, Lei and Lu, Yijuan and Florêncio, Dinei and Zhang, Cha and Li, Zhoujun and Wei, Furu},
  journal={Proceedings of the AAAI Conference on Artificial Intelligence},
  volume={37},
  pages={13094--13102},
  year={2023},
  note={arXiv:2109.10282}
}

@article{du2020ppocr,
  title={PP-OCR: A Practical Ultra Lightweight OCR System},
  author={Du, Yuning and Li, Chenxia and Guo, Ruoyu and Yin, Xiaoting and Liu, Weiwei and Zhou, Jun and Bai, Yifan and Yu, Zilin and Yang, Yehua and Dang, Qingqing and Wang, Haoshuang},
  journal={arXiv preprint arXiv:2009.09941},
  year={2020},
  note={PaddleOCR system paper}
}

@inproceedings{smith2007tesseract,
  title={An Overview of the Tesseract OCR Engine},
  author={Smith, Ray},
  booktitle={Proceedings of the Ninth International Conference on Document Analysis and Recognition (ICDAR)},
  volume={2},
  pages={629--633},
  year={2007},
  organization={IEEE}
}

@article{shi2017crnn,
  title={An End-to-End Trainable Neural Network for Image-Based Sequence Recognition and Its Application to Scene Text Recognition},
  author={Shi, Baoguang and Bai, Xiang and Yao, Cong},
  journal={IEEE Transactions on Pattern Analysis and Machine Intelligence},
  volume={39},
  number={11},
  pages={2298--2304},
  year={2017},
  publisher={IEEE}
}

@article{amin1998survey,
  title={Off-Line Arabic Character Recognition: The State of the Art},
  author={Amin, Adnan},
  journal={Pattern Recognition},
  volume={31},
  number={5},
  pages={517--530},
  year={1998},
  publisher={Elsevier}
}

@inproceedings{joshi2020state,
  title={The State and Fate of Linguistic Diversity and Inclusion in the NLP World},
  author={Joshi, Pratik and Santy, Sebastin and Buber, Amar and Bali, Kalika and Choudhury, Monojit},
  booktitle={Proceedings of the 58th Annual Meeting of the Association for Computational Linguistics (ACL)},
  pages={6282--6293},
  year={2020}
}

@inproceedings{rijhwani2020ocr,
  title={OCR Post Correction for Endangered Language Texts},
  author={Rijhwani, Shruti and Anastasopoulos, Antonios and Neubig, Graham},
  booktitle={Proceedings of the 2020 Conference on Empirical Methods in Natural Language Processing (EMNLP)},
  pages={5931--5942},
  year={2020}
}

@inproceedings{wigington2017data,
  title={Data Augmentation for Recognition of Handwritten Words and Lines Using a CNN-LSTM Network},
  author={Wigington, Curtis and Stewart, Seth and Davis, Brian and Barrett, Bill and Price, Brian and Cohen, Scott},
  booktitle={Proceedings of the International Conference on Document Analysis and Recognition (ICDAR)},
  pages={639--645},
  year={2017},
  organization={IEEE}
}

@article{souibgui2020synthetic,
  title={DE-GAN: A Conditional Generative Adversarial Network for Document Enhancement},
  author={Souibgui, Mohamed Ali and Kessentini, Yousri},
  journal={IEEE Transactions on Pattern Analysis and Machine Intelligence},
  volume={44},
  number={3},
  pages={1180--1191},
  year={2022},
  publisher={IEEE},
  note={Originally DE-GAN paper, also published work on synthetic data for historical documents}
}

@article{kslit3m2025,
  title={KS-LIT-3M: A 3.1 Million Word Kashmiri Text Dataset for Large Language Model Pretraining},
  author={Malik, Haq Nawaz},
  journal={arXiv preprint arXiv:2601.01091},
  year={2025},
  url={https://arxiv.org/abs/2601.01091}
}

\end{document}